\documentclass[10pt,twocolumn,letterpaper]{article}
\PassOptionsToPackage{table,dvipsnames}{xcolor}
\usepackage{iccv}
\usepackage{cuted}
\usepackage{caption}
\usepackage{amsmath,amssymb}
\usepackage{booktabs,multirow,makecell}
\usepackage{xcolor,colortbl}
\usepackage{graphicx}
\usepackage{enumitem}
\usepackage{xspace}
\usepackage{tikz}

\newcommand{\ours}{ArtLang-4D\xspace}
\DeclareMathOperator*{\argmax}{arg\,max}
\newcommand{\norm}{\operatorname{n}}
\providecommand{\SE}{\mathrm{SE}(3)}
\definecolor{ourscolor}{RGB}{220,230,240}

\definecolor{cArt}{HTML}{1F7A74}
\definecolor{cLang}{HTML}{6F5BA8}
\definecolor{cTime}{HTML}{C9841C}
\newcommand{\mk}[2]{\tikz[baseline=(m.base)]{\node[circle,draw=#2!80!black,fill=#2!20!white,line width=0.45pt,inner sep=0pt,minimum size=1.3em,font=\sffamily\bfseries\scriptsize,text=#2!50!black](m){#1};}}
\newcommand{\mA}{\mk{A}{cArt}\,}
\newcommand{\mL}{\mk{L}{cLang}\,}
\newcommand{\mT}{\mk{T}{cTime}\,}
\newcommand{\mkw}[2]{\tikz[baseline=(m.base)]{\node[rounded corners=2.5pt,draw=#2!80!black,fill=#2!20!white,line width=0.45pt,inner xsep=2.2pt,inner ysep=0pt,minimum height=1.3em,font=\sffamily\bfseries\scriptsize,text=#2!50!black](m){#1};}}
\newcommand{\mAG}{\mkw{A-Geo}{cArt}\,}
\newcommand{\mAP}{\mkw{A-Part}{cArt}\,}
\newcommand{\mAA}{\mkw{A-Joint}{cArt}\,}

\definecolor{iccvblue}{rgb}{0.21,0.49,0.74}
\usepackage[pagebackref,breaklinks,colorlinks,allcolors=iccvblue]{hyperref}

\def\paperID{*****}
\def\confName{ICCV}
\def\confYear{2026}

\title{ArtLang-4D: Articulation-Aware Dynamic Language Fields from Monocular Videos}

\author{
\textbf{Sylvia Yuan}\textsuperscript{1,*},
\textbf{Dan Wang}\textsuperscript{2,*},
\textbf{Ravi Ramamoorthi}\textsuperscript{1},
\textbf{Xinrui Cui}\textsuperscript{3,\dag}
\\\\
\textsuperscript{1}University of California San Diego \quad
\textsuperscript{2}University of Arizona \quad
\textsuperscript{3}University of North Texas \\
\\
{\tt\small \textsuperscript{*}Equal Contribution \quad \textsuperscript{\dag}Corresponding Author}
}

\begin{document}
\maketitle
\begin{strip}
\centering
\includegraphics[width=0.84\linewidth]{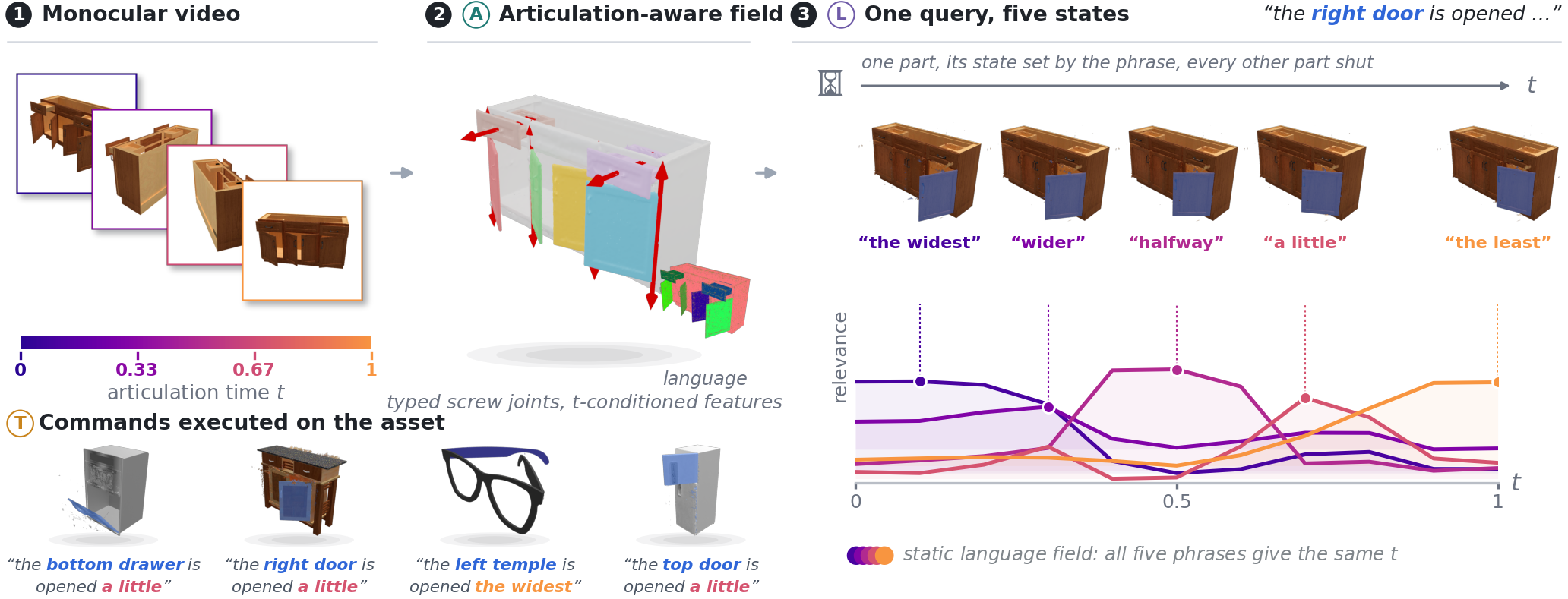}
\captionsetup{hypcap=false}
\captionof{figure}{\textbf{\ours turns one video of an articulated object into a model that understands state-dependent language.} From a monocular RGB-D video in which the camera moves and the parts move, we recover a canonical triangle field with typed joints, \protect\mA, and a language field conditioned on the articulation state $t$, \protect\mL. Five phrases about the same door peak at five different states, and each render shows that door alone posed at the state its phrase selects, every other part shut; a static language field gives all five the same answer. A command moves only the selected part along its recovered joint, \protect\mT. The part word of each query is tinted with the moved part's color, and the state word with the color of $t$.}
\captionsetup{hypcap=true}
\label{fig:teaser}
\end{strip}

\begin{abstract}
Language that describes an articulated object is often about its state: \emph{the drawer pulled out the furthest}, \emph{the door opened halfway}, \emph{push the drawer all the way in}. Existing 3D language fields attach one feature to each point and cannot tell these descriptions apart, and 4D language fields condition on video time, which has no kinematic meaning and cannot be set by a user. We present \ours, which reconstructs an articulated object from a single monocular RGB-D video and equips it with a dynamic language field indexed by articulation time $t$, the normalized state of its joints, instead of video time. The geometry is a canonical triangle field whose triangles follow typed screw joints, discovered from the video without any joint initialization. The language field is supervised by a $t$-tagged caption bank that describes each part with comparative state phrases in a whitened open-vocabulary text space. A free-form query is answered by maximizing part relevance jointly over parts and states, which returns both the part and the articulation state it describes; a command is then executed by moving only that part along its recovered joint. On the 100 objects of Articulate-100, \ours recovers more accurate joints and part geometry than reconstruction and articulation baselines, retrieves the queried state far more reliably than a state-agnostic language field, and executes language commands with lower motion error.
\end{abstract}

\section{Introduction}
\label{sec:intro}

Interacting with articulated objects through language requires understanding both which part to move and where along its motion to place it. ``Move the left drawer halfway between its observed positions'' identifies a persistent component and a target configuration. Answering this request requires \textit{geometry}, \textit{semantics}, and \textit{kinematics} to agree: the system must \textit{recognize the drawer}, \textit{interpret its state relative to a motion interval}, and \textit{execute the corresponding joint displacement}. We study how to establish this correspondence on an articulated asset reconstructed from video.

Existing methods provide complementary foundations. Open-vocabulary 3D language fields~\cite{lerf,langsplat,openscene} associate semantic features with scene geometry, while articulated reconstruction methods~\cite{paris,artgs,gaussianart,artmesh,videoartgs} recover parts and joints. These advances motivate a more challenging question: \textit{how can a dynamic semantic field directly represent an articulated state description in the time-varying joint coordinates used to execute it?} A recording timestamp identifies an observation, whereas a dynamic joint coordinate specifies an articulated configuration of a persistent part. Connecting semantic retrieval to that coordinate provides an explicit interface between describing an articulated state and setting it on the reconstructed object.

Establishing this correspondence would make captured objects directly addressable through descriptions of their configurations, supporting interactive reconstruction, simulation, and digital asset editing. This problem presents three challenges. \mA \textit{Recovering persistent \underline{\textbf{A}}rticulation identity under motion:} camera and part motion change the observed image, while the surface associated with a named part must remain consistent. \mL \textit{State-dependent \underline{\textbf{L}}anguage:} the same part must match different descriptions at different coordinates, with comparative expressions interpreted relative to an observed motion range. \mT \textit{Executable \underline{\textbf{T}}emporal grounding:} the selected state must correspond to the recovered joint motion, and controlling one part must preserve the current coordinates of the others. Consequently, semantic retrieval, geometric reconstruction, and joint estimation must be considered together.  

We introduce \ours, a dynamic 4D articulated language-embedded triangle field that couples \mA persistent articulated triangle-based geometry, part identity, joint motion, \mL state-conditioned language features, and \mT temporal querying and control. The central idea is to use a shared 4D articulation coordinate for \textit{semantic language evaluation and geometric articulation execution}. A triangle primitive retains its part identity as the asset moves; its dynamic language feature changes with the coordinate used by that part’s revolute or prismatic transform. This makes state retrieval an explicit search over controllable configurations of the reconstructed asset. 

Our contributions are:
\begin{itemize}[nosep,leftmargin=*]
\item \mA\textbf{Persistent articulation identity from one monocular video.} A canonical triangle field whose part identity field comes from $t$-free spatial planes, so a triangle keeps its part as camera and parts move, with a joint transform field of typed screw joints discovered from 3D part observations without initialization and guarded against photometric drift (Secs.~\ref{sec:artic} and~\ref{sec:exp_art}).
\item \mL\textbf{State-dependent language in joint coordinates.} A state-dependent language field, a per-triangle feature modulated by the articulation state and supervised by a $t$-tagged bank of comparative captions in a whitened text space, so one part matches different descriptions at different states (Secs.~\ref{sec:language} and~\ref{sec:exp_lang}).
\item \mT\textbf{Executable temporal grounding.} A query that maximizes relevance over parts and states together and returns a pair $(k^*,t^*)$ executed on that part alone, with type-aware naming of the parts (Secs.~\ref{sec:control} and~\ref{sec:exp_ctrl}).
\end{itemize}
On Articulate-100 rendered as monocular articulation videos, \ours improves over reconstruction and articulation baselines on every motion, geometry and language metric, and the same pipeline runs on real phone captures (Sec.~\ref{sec:exp}).

\section{Related Work}
\label{sec:related}

\textbf{Articulated object reconstruction.} \mA Implicit methods reconstruct articulated objects from two static states, from point clouds before and after an interaction~\cite{ditto} or multi-view images~\cite{paris,digitaltwinart}, as do Gaussian-splatting methods~\cite{artgs,gaussianart,splart,reartgspp}, and ArtMesh~\cite{artmesh} recovers part-aware meshes with motion-consistent joints in the same setting. Fewer methods start from video. iTACO~\cite{video2articulation} predicts movable parts and joints from a scan and an interaction video, and VideoArtGS~\cite{videoartgs} reconstructs Gaussians and joints from a monocular video using 3D tracks as a motion prior. Feedforward and vision-language approaches estimate or generate articulation from images or object assets~\cite{opd,real2code,singapo,articulate-anything}. None of these attach language to the recovered parts. We reconstruct from a single monocular video as VideoArtGS does, but with an explicit canonical triangle field and screw joints that are discovered without seeds, so that the joint state can index a language field.

\textbf{Language fields.} \mL OpenScene~\cite{openscene}, LERF~\cite{lerf} and LangSplat~\cite{langsplat} distill vision-language features onto point clouds, radiance fields and Gaussians, Feature 3DGS~\cite{feature3dgs} extends the idea to segmentation and editing, and GARField~\cite{garfield} groups scenes from masks without language. These fields describe static scenes. 4D LangSplat~\cite{4dlangsplat} builds time-sensitive features from multimodal-LLM captions of tracked objects and deforms them with a HexPlane~\cite{hexplane} over video time, on top of a free deformation of 4D Gaussians~\cite{4dgs}. Our field is conditioned on an articulation state instead. The state has a kinematic meaning shared by every part, it can be set by a user, and the same coordinate drives the rigid motion of the geometry.

\textbf{Language-driven articulation.} \mT Articulate-Anything~\cite{articulate-anything} and URDF-Anything~\cite{urdfanything} use vision-language models to produce articulated assets, A3VLM~\cite{a3vlm} predicts part boxes, joint axes and joint types from one image, and video generation models animate parts from drags~\cite{puppetmaster} or text~\cite{atop}. ArtLang~\cite{artlang} grounds commands to the parts and joints of a reconstruction, with a static semantic field and two observed states. \ours differs in two ways. Its part semantics change with the state, so the state itself can be retrieved from language, and it is learned from one monocular video rather than two multi-view captures.

\section{Method}
\label{sec:method}

\begin{figure*}[t]
\centering
\includegraphics[width=0.88\textwidth]{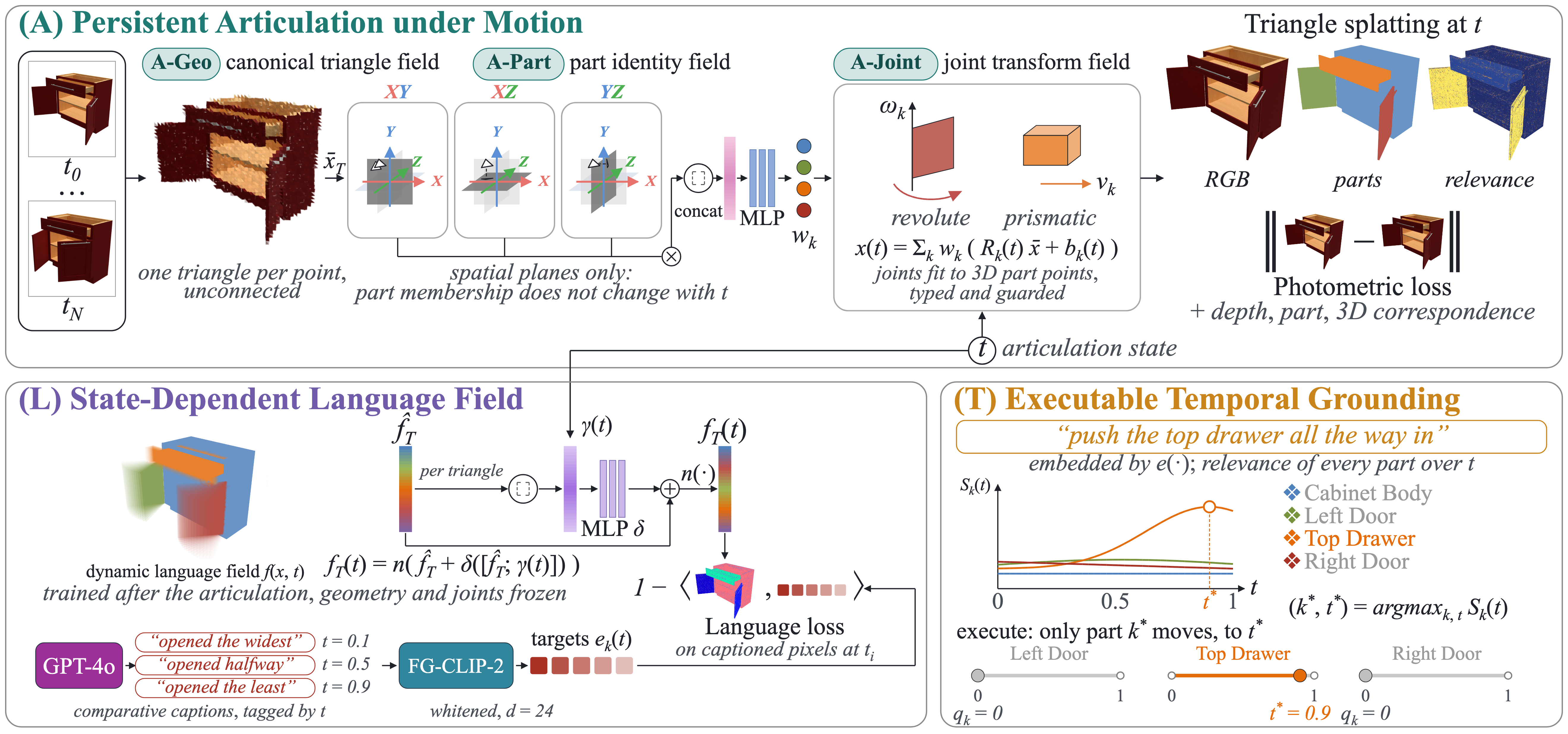}
\caption{\textbf{Overview of \ours.} \protect\mA A canonical triangle field is fit to the video (A-Geo); a part identity field decoded from $t$-free spatial planes gives the weights $w_k$, so membership does not change with $t$ (A-Part); a joint transform field moves every part by a typed screw joint discovered from 3D part observations (A-Joint), Eq.~\eqref{eq:blend}; and the field is splatted at $t$ into RGB, part and language maps under a photometric loss. \protect\mL A state-dependent language field: a per-triangle feature $\hat f_T$ modulated by $\gamma(t)$, Eq.~\eqref{eq:lang}, trained with the geometry frozen against $t$-tagged caption targets $e_k(t)$. \protect\mT A query is embedded in the same space and the maximizing $(k^*,t^*)$ is executed on part $k^*$ alone.}
\label{fig:framework}
\end{figure*}

\textbf{Problem formulation.} Our objective is to ground a target-state description to a geometric part and an articulated configuration that can be executed on a reconstructed 3D asset. The key design is to evaluate dynamic surface semantics in the same 4D articulation coordinate that drives typed geometric motion.
We consider an object with a fixed base, indexed by $0$, and $K$ movable parts. Each movable part is attached directly to the base by one revolute or prismatic joint. The input is a monocular video observation $\mathcal D=\{I_i,D_i,\Pi_i\}_{i=1}^N$, with images $I_i$, depth maps $D_i$, cameras $\Pi_i$. The video is RGB-D: $D_i$ is rendered for the synthetic videos and recorded by the phone's depth sensor for the real captures (Sec.~\ref{sec:exp_real}). An articulation state $t\in[0,1]$ runs between the two observed reference states of the object; its zero state defines the canonical geometry and is not assumed to be mechanically closed, and frame $i$ observes the object at state $t_i$.

\textbf{Overview.} Figure~\ref{fig:framework} summarizes the pipeline, and the next three sections follow its three panels: the articulated surface field \mA (Sec.~\ref{sec:artic}), the language field \mL (Sec.~\ref{sec:language}) and the temporal language querying and articulation grounding \mT (Sec.~\ref{sec:control}).

\mA \mL \textbf{Dynamic 4D articulated language-embedded triangle field}. We reconstruct the canonical 3D object as a canonical triangle field \(\mathcal{T}\) \mAG~\cite{artmesh,triangle-splatting}. For a point $x$ on canonical triangle primitive $T$, we learn our dynamic 4D articulated language-embedded triangle field as
\begin{equation}
\Phi(x,t)=
 \big(\pi(x),G_{\pi(x)}(t),f(x,t)\big),
\label{eq:art4d_field}
\end{equation}
where $\pi(x)$ is a fixed state-independent part identity field \mAP, $G_k(t)\in\SE$ a part-wise joint transform field \mAA, and $f(x,t)$ a state-dependent language field \mL.

\mT \textbf{Temporal querying and grounding.} For a language query \(s\), let \(e(s)\) be its normalized text embedding, let \(\mathcal{T}_k\) contain the triangle primitives assigned to part \(k\), and let \(f(x_T,t)\) be triangle \(T\)’s normalized dynamic language feature at candidate state \(t\). Grounding evaluates
\begin{equation}\begin{aligned} S_k(t;s) &= \frac{1}{|\mathcal{T}_k|} \sum_{T\in\mathcal{T}_k} \left\langle f(x_T,t),e(s)\right\rangle,\\ (k^\star,t^\star) &= \underset{k,\;t\in\mathcal{G}}{\operatorname{arg\,max}} \,S_k(t;s), \label{eq:query_grounding} \end{aligned} \end{equation}
where \(\mathcal{G}\) is the state-search grid. The selected coordinate \(t^\star\) is passed to part \(k^\star\)’s recovered transform; every other part retains its current coordinate. The retrieved state thus has an explicit geometric interpretation.

\subsection{Persistent Articulation under Motion}
\label{sec:artic}
\label{sec:soup}
\label{sec:joints}

\mA The part a point belongs to, and the motion that part makes, have to survive the motion of the camera and of the parts. Three fields, the top row of Figure~\ref{fig:framework}, keep them fixed: a canonical triangle field (A-Geo), a part identity field that does not depend on the state (A-Part), and a joint transform field with one typed joint per part, discovered without initialization (A-Joint).

\mAG\textbf{Canonical triangle field.} We represent the object as a canonical triangle field: a set of unconnected triangles, the primitive of triangle splatting~\cite{triangle-splatting} and the initial state of the mesh field of ArtMesh~\cite{artmesh}. Each triangle $T$ stores three vertices with positions $x_v$, spherical-harmonic colors and opacities $o_v$. A differentiable rasterizer gives $T$ a soft footprint at pixel $p$ from the signed distance $\mathrm{sd}_T(p)$ to its projected edges,
\begin{equation}
  \alpha_T(p)=o_T(p)\Bigl(\max\bigl(0,\,\mathrm{sd}_T(p)/\mathrm{sd}_T^{\max}\bigr)\Bigr)^{\sigma},
  \label{eq:window}
\end{equation}
which is $1$ farthest from the edges, $0$ on them and sharpens with $\sigma$, with $o_T(p)$ interpolated from the vertex opacities. Triangles are alpha-composited front to back,
\begin{equation}
  C(p)=\sum_{T}c_T(p)\,\alpha_T(p)\prod_{T'\prec T}\bigl(1-\alpha_{T'}(p)\bigr),
  \label{eq:composite}
\end{equation}
and the same sum splats the part weights and language features below. The field has no connectivity, so a triangle never spans two parts and every triangle can move rigidly with its own part. ArtMesh then remeshes each part into a connected explicit mesh by part-aware restricted Delaunay remeshing. We do not, because the triangles have to stay learnable primitives throughout: vertices, opacities, part weights and language features are updated, and triangles densified and pruned, until the end of training. The field is initialized with one triangle per point of a cloud fused from the video's depth, densified by subdividing high-importance triangles, and pruned when a triangle stays transparent or never contributes to a pixel, following the schedule of ArtMesh~\cite{artmesh}.

\mAP\textbf{Part identity field.} Let $\bar x_T$ be the centroid of triangle $T$. A small decoder $h$ reads multi-resolution spatial feature planes $\varphi$ (the $xy$, $xz$ and $yz$ planes of a HexPlane~\cite{hexplane}, with no state dependence) and outputs part weights
\begin{equation}
  \mathbf w_T=\operatorname{softmax}\bigl(h(\varphi(\bar x_T))\bigr)\in\Delta^{K}.
  \label{eq:weights}
\end{equation}
Decoding at the centroid gives all three vertices of a triangle the same weights. Because $\varphi$ ignores $t$, a triangle belongs to the same part at every state. After training, with geometry and joints fixed, free per-triangle logits initialized from the decoder are refined with the splatted part loss of Eq.~\eqref{eq:loss}, which sharpens labels the smooth decoder blurs across part boundaries.

\mAA\textbf{Joint transform field.} Each movable part $k\ge1$ has a rotation vector $\boldsymbol\omega_k\in\mathbb R^3$, a translation vector $\mathbf v_k\in\mathbb R^3$, a pivot $\mathbf p_k\in\mathbb R^3$, and a type given by two indicators $r_k,s_k\in\{0,1\}$ with $r_k+s_k=1$ after type selection. Its motion at state $t$ is
\begin{equation}
\begin{aligned}
  R_k(t)&=\exp\!\bigl(t\,r_k[\boldsymbol\omega_k]_\times\bigr),\\
  b_k(t)&=\mathbf p_k-R_k(t)\,\mathbf p_k+t\,s_k\mathbf v_k,
\end{aligned}
  \label{eq:screw}
\end{equation}
so a revolute joint ($r_k{=}1$) rotates by the angle $t\|\boldsymbol\omega_k\|$ about the axis $\boldsymbol\omega_k/\|\boldsymbol\omega_k\|$ through $\mathbf p_k$, and a prismatic joint ($s_k{=}1$) slides by $t\mathbf v_k$. The base is fixed, $R_0=I$ and $b_0=\mathbf 0$. A vertex $x$ of triangle $T$ moves to
\begin{equation}
  x(t)=\sum_{k=0}^{K}w_{T,k}\bigl(R_k(t)\,x+b_k(t)\bigr).
  \label{eq:blend}
\end{equation}
During training the weights are soft, so the part boundaries remain differentiable. For export and control each triangle follows its most likely part $\pi(T)=\argmax_k w_{T,k}$, and Eq.~\eqref{eq:blend} reduces to the joint transform $G_{\pi(T)}(t)=(R_{\pi(T)}(t),b_{\pi(T)}(t))$ of Eq.~\eqref{eq:art4d_field}. Unlike a free deformation field~\cite{4dgs,4dlangsplat}, Eq.~\eqref{eq:blend} gives $t$ the same kinematic meaning for every triangle, and setting $t$ for one part moves that part alone.

\mAA\textbf{Seed-free joint discovery.} Photometric losses constrain joints only weakly when the camera and the parts move together, so we first discover the joints from 3D evidence and then train them with the triangle field, the A-Joint block of Figure~\ref{fig:framework}.

\emph{3D part observations.} For every frame $i$ and part $k$, we erode the mask region of $k$, drop pixels at depth discontinuities, and back-project the remaining pixels with $D_i$ and $\Pi_i$ into a point set $Y_k^i$. The frames at the rest state, $t_i=0$, give the canonical points $X_k=\bigcup_{i:t_i=0}Y_k^i$.

\emph{Robust correspondence.} A joint explains frame $i$ when the canonical points of its part, moved by Eq.~\eqref{eq:screw}, cover the observed points. With $\ell$ the diagonal of the object's bounding box and the Cauchy kernel $\rho_\sigma(d)=\sigma^2\log(1+d^2/\sigma^2)$, we define
\begin{equation}
  \mathcal L_{\rm corr}^{i}=\frac{1}{\ell^2|\mathcal K|}\sum_{k\in\mathcal K}\frac{1}{|Y_k^i|}\sum_{y\in Y_k^i}\rho_\sigma\Bigl(\min_{x\in X_k}\bigl\|y-R_k(t_i)x-b_k(t_i)\bigr\|\Bigr),
  \label{eq:corr}
\end{equation}
where $\mathcal K$ is the set of movable parts visible in frame $i$. The kernel behaves like a squared distance for small residuals and grows only logarithmically for large ones. This matters because an opening part reveals surfaces that $X_k$ does not contain, such as the side walls of a drawer, and a non-robust loss would drag the part towards them. We anneal $\sigma$ geometrically from $0.07\ell$ to $0.0035\ell$, so the fit first finds a coarse basin and then tightens. During training, $X_k$ is replaced by the centroids of the triangles currently assigned to part $k$, and the loss updates only the joint parameters.

\emph{Type selection and multi-start fitting.} We fit $(\boldsymbol\omega_k,\mathbf v_k,\mathbf p_k)$ with both indicators on, $r_k=s_k=1$, so that Eq.~\eqref{eq:screw} is a general screw motion that rotates and slides at once, by minimizing Eq.~\eqref{eq:corr} over random frames with a curriculum that admits states up to $\min(1,2\tau)$ at normalized step $\tau$. We then decide the type. From the free fit we build two revolute hypotheses (the screw axis of the fitted motion, and a rotation that starts at zero about the part centroid) and one prismatic hypothesis (the displacement of the part centroid), refit each hypothesis, and choose revolute only if
\begin{equation}
  \min\bigl(\mathcal L_{\rm rev}^{a},\mathcal L_{\rm rev}^{b}\bigr)<(1-m)\,\mathcal L_{\rm pri},
  \label{eq:type}
\end{equation}
with margin $m=0.1$, since a revolute joint can mimic a short slide while the converse does not hold. The chosen type is locked and the joint is re-optimized under that type alone, with the other indicator set to zero, for a further 2{,}000 steps. We repeat this from $S=6$ starts, one at the default initialization and five at random axes and pivots inside the part's bounding box, take the majority type per part, and keep the start of that type with the lowest observation fit $E_k$, which is the mean of the part's term in Eq.~\eqref{eq:corr} over all moving frames at the final $\sigma$.

\mAA\textbf{Training and joint guard.} The triangle field, the decoder and the joints are then trained together by
\begin{equation}
\begin{aligned}
  \mathcal L={}&\mathcal L_{\rm rgb}+\lambda_d\mathcal L_{\rm depth}+\lambda_o\mathcal L_{\rm occ}\\
  &+\lambda_p\mathcal L_{\rm part}+\lambda_c\mathcal L_{\rm corr}+\mathcal L_{\rm reg}.
\end{aligned}
  \label{eq:loss}
\end{equation}
$\mathcal L_{\rm rgb}=0.8\|\hat I_i-I_i\|_1+0.2\,(1-\mathrm{SSIM}(\hat I_i,I_i))$ compares the render of the triangle field posed at $t_i$ with the frame. $\mathcal L_{\rm depth}$ is the $L_1$ error of the rendered depth, normalized by the camera extent, and $\mathcal L_{\rm occ}=-\log(1-\hat\alpha)$ on background pixels discourages opacity outside the object. $\mathcal L_{\rm part}$ splats the weights $\mathbf w_T$ into an image, normalizes them per pixel, and applies a cross-entropy against $M_i$. $\mathcal L_{\rm reg}$ collects the field's opacity and normal regularizers, an entropy term on $\mathbf w_T$, and a smoothness term on the feature planes. We use $\lambda_d=0.2$, $\lambda_o=0.5$, $\lambda_p=0.05$ and $\lambda_c=1$. Training starts with a coarse stage on the rest frames only, with no motion, and then admits frames in order of increasing $t$, using the same curriculum as discovery. Photometric gradients can still move a joint away from what the 3D evidence supports, so after training we compare the fit $E_k$ of the trained and discovered joints and revert joint $k$ to the discovered one if $E_k^{\rm trained}>1.25\,E_k^{\rm disc}$.

\subsection{State-Dependent Language Field}
\label{sec:language}

\mL The same part has to match different descriptions at different states, so every triangle carries a language feature that is a function of $t$, supervised by captions tagged with the state they describe (Figure~\ref{fig:framework}, lower left).

\textbf{$t$-tagged caption bank.} The language targets describe each part at several states with phrases that compare states. GPT-4o~\cite{gpt4o} receives two images of the object at the two ends of its motion, taken from the same viewpoint (in our synthetic experiments, renders at $t=0$ and $t=1$). It lists the movable parts with their joint types and states which end is opened wider. It is asked twice with the two images in opposite orders. A consistent answer is kept, and when the answers disagree, the end in which the parts cover a larger image area is used. The names are matched to the mask labels by Hungarian assignment on image centroids. We split $[0,1]$ into $B=5$ buckets with centers $c_b$, and for each movable part and bucket we compute the openness $\eta_b$, equal to $c_b$ when the part opens wider at $t=1$ and to $1-c_b$ otherwise. The caption is \emph{``the \{part\} is \{phrase\}''}, where the phrase is entry $\min(4,\lfloor5\eta_b\rfloor)$ of a five-step scale chosen by the joint, such as \emph{opened the least}, \emph{opened a little}, \emph{opened halfway between its two positions}, \emph{opened wider} and \emph{opened the widest} for a swinging part, with \emph{pulled out} in place of \emph{opened} for a sliding part. The words are comparative because the two observed states of a part are often both partly open. The base receives its name alone.

\textbf{Whitened text space.} Every caption $s$ is embedded by the FG-CLIP~2 text encoder $\psi$~\cite{fgclip2}. Raw CLIP-style embeddings of short part phrases are dominated by a few shared directions, so \emph{left door} and \emph{right door} lie almost on top of each other when projected without whitening, and the five state phrases of Figure~\ref{fig:teaser} separate only after this step. We therefore fit a whitened PCA on the captions of the object together with a fixed vocabulary of part and state phrases, with mean $\mu$ and the top $d=24$ principal directions $V_d$ scaled by their inverse singular values $\Sigma_d^{-1}$, and map
\begin{equation}
  e(s)=\norm\bigl((\psi(s)-\mu)\,V_d\Sigma_d^{-1}\bigr),
  \label{eq:embed}
\end{equation}
where $\norm(\cdot)$ denotes $L_2$ normalization. The caption of part $k$ in bucket $b$ gives the target $e_{k,b}$, and a frame at state $t_i$ uses the bucket whose center is nearest, written $e_k(t_i)$. The same $\psi$, $\mu$ and $V_d\Sigma_d^{-1}$ embed every query at test time, so any phrase can be compared with the field.

\textbf{$t$-conditioned feature.} Each triangle stores a feature $f_T\in\mathbb R^d$. A small MLP $\delta$ modulates its normalized value $\hat f_T=\norm(f_T)$ by the state,
\begin{equation}
\begin{aligned}
  f_T(t)&=\norm\bigl(\hat f_T+\delta([\hat f_T;\gamma(t)])\bigr),\\
  \gamma(t)&=\bigl[t,\{\sin 2^jt,\cos 2^jt\}_{j=0}^{3}\bigr].
\end{aligned}
  \label{eq:lang}
\end{equation}
The static part $\hat f_T$ carries what the triangle is, and the residual carries how the description changes with the state. We pose the triangle field at $t_i$, splat $f_T(t_i)$ through the same rasterizer into a feature image $F$, and minimize
\begin{equation}
  \mathcal L_{\rm lang}=\frac{1}{|\Omega_i|}\sum_{p\in\Omega_i}\Bigl(1-\bigl\langle\norm(F_p),\,e_{M_i(p)}(t_i)\bigr\rangle\Bigr),
  \label{eq:langloss}
\end{equation}
where $\Omega_i$ holds the pixels of frame $i$ that belong to a captioned part. In this stage the triangle field, the part decoder and the joints are frozen, so language cannot change the geometry or the motion. The output is the complete field of Eq.~\eqref{eq:art4d_field}: every triangle carries its part, its typed joint and a language feature $f_T(t)$ that changes with the state (Figure~\ref{fig:framework}, L; the curves of Figure~\ref{fig:teaser} are its relevance over $t$).

\subsection{Executable Temporal Grounding}
\label{sec:control}

\mT The state a description selects has to be one the recovered joints can realize. Since part, motion and language feature share $t$ in Eq.~\eqref{eq:art4d_field}, a query is a search over parts and states and its answer is a joint configuration (Figure~\ref{fig:framework}, lower right).

\textbf{Part and state from one query.} Eq.~\eqref{eq:query_grounding} scores part $k$ at state $t$ by the mean similarity of its features $f_T(t)$ to the query embedding $e(s)$, over $\mathcal T_k=\{T:\pi(T)=k\}$, and searches $\mathcal G$, a uniform grid of 11 states on $[0,1]$. The maximum over $t$ is what makes the query temporal; a static field has a flat $S_k(t;s)$ and can name a part but not a state. The lower right of Figure~\ref{fig:framework} traces one query to the top drawer at $t^*=0.9$, where the field describes it as pulled out the least.

\textbf{Executing a command.} The per-part configuration $\mathbf q$ sets $q_{k^*}=t^*$ and keeps $q_k=q_k^{\rm cur}$ for every other part ($0$ for a freshly loaded asset); each triangle is posed by $G_{\pi(T)}(q_{\pi(T)})$. When the query selects the base, no joint moves.

\textbf{Naming parts.} Retrieving \emph{the left door} needs a name per part. We match the GPT-4o names $n_j$ to the parts by a Hungarian assignment with cost
\begin{equation}
\begin{aligned}
  C_{kj}={}&\bigl(1-\langle\mathbf a_k,e(n_j)\rangle\bigr)+\|\mathbf c_k-\mathbf z_j\|_2\\
  &+2\,[\,\kappa_k\neq\kappa_j\,],
\end{aligned}
  \label{eq:naming}
\end{equation}
where $\mathbf a_k=\norm\bigl(\sum_{t\in\mathcal G}\frac{1}{|\mathcal T_k|}\sum_{T\in\mathcal T_k}f_T(t)\bigr)$ is the language prototype of part $k$, $\mathbf c_k$ and $\mathbf z_j$ are the image centroid of part $k$ and the center of the box GPT-4o drew for $n_j$, in normalized coordinates of the same rest view, and the last term charges a disagreement between the recovered joint type $\kappa_k$ and the type $\kappa_j$ GPT-4o proposed. The base is pinned to the fixed part. We call this type-aware naming.

\section{Experiments}
\label{sec:exp}

\subsection{Setup}
\label{sec:setup}

\textbf{Data.} The main evaluation is on rendered videos, where every joint, part and state has ground truth; Sec.~\ref{sec:exp_real} runs the same pipeline on real captures. We evaluate on Articulate-100~\cite{artmesh}, 100 PartNet-Mobility~\cite{sapien} objects across 16 categories with two to more than six parts. Each object is rendered as one monocular video of 172 frames at $512\times512$: the camera orbits and spirals around the object at rest for 100 frames, then keeps orbiting for two more turns while $t$ rises linearly to $1$ over 72 frames, so every part is seen moving from several directions. Each frame has its camera, depth and state $t$, and $t=0$ and $t=1$ are the two states of Articulate-100. Part masks can come from two sources, the ground truth rendered with each frame or the predictions of a segmentation model such as SAM~3~\cite{sam3} or the finetuned PArtSAM of ArtMesh~\cite{partsam,artmesh}.

\textbf{Baselines.} VideoArtGS~\cite{videoartgs} reconstructs articulated Gaussians from a monocular video, taking posed RGB-D frames with a foreground mask, and is the closest baseline in input. It receives the same frames, depth and cameras, the state $t$ as its time variable, and the joint count and types from the same GPT-4o step that we use, in place of its own vision-language query. ArtLang~\cite{artlang} reconstructs from multi-view images of the two end states, which is a stronger input than one video, and it is the only baseline with a language interface. Articulate-Anything~\cite{articulate-anything} receives the stretch of the same video in which the parts move, as RGB frames only, retrieves a PartNet-Mobility template and predicts the joints of that template with GPT-4o in place of its default Gemini, scored under the better of its two state orderings (Sec.~\ref{sec:supp_baselines} in the appendix). Following iTACO~\cite{video2articulation}, an object on which a method fails scores $90^\circ$ for axis angle and revolute motion and $1$\,m for every other error, and we extend the rule per part: a ground-truth part that a method does not predict scores the same penalty on that part only.

\textbf{Metrics.} For motion we report the angle between predicted and true joint axes, the distance between axis lines for revolute joints (zero for prismatic joints, as in the released evaluation of ArtGS~\cite{artgs}), and the part motion error at $t=1$. For geometry we report the Chamfer distance on static and movable parts with 10{,}000 samples per mesh, averaged over the two end states. For language we use the protocol of ArtLang~\cite{artlang}: LSE-A and LSP-A measure, at held-out cameras in both end states, the IoU between the parts that carry a given noun or spatial word and the ground-truth parts with the same word; LM-A pairs predicted and true parts by name, moves both by $\alpha\in\{-1,-\tfrac12,0,\tfrac12,1\}$ of their joint and averages the Chamfer distance. State Acc is the fraction of parts, over the frames of the input video, whose rendered feature matches the caption of their own state best among the five state captions.

\textbf{Implementation.} Joint discovery, triangle field training, the language stage and label refinement take about two hours per object on one GPU; Sec.~\ref{sec:supp_impl} in the appendix lists every hyperparameter and iteration count, and Secs.~\ref{sec:supp_metrics} and~\ref{sec:supp_baselines} the full metric and baseline protocols.

\begin{table*}[t]
\centering
\caption{\textbf{Comparison on Articulate-100} (100 objects). Motion: axis angle ($^\circ$), axis position (0.1\,m), part motion ($^\circ$ or m). Geometry: Chamfer distance on static and movable parts (mm). Language: commanded-motion error LM-A (mm), semantic and spatial part retrieval LSE-A and LSP-A (IoU), and state retrieval accuracy over five states. Failed objects and missing parts receive the iTACO penalty~\cite{video2articulation}. Language control marks methods that execute part commands given in language. Best in \textbf{bold}, second \underline{underlined}. Gain is the improvement over the best baseline in each column.}
\label{tab:main}
\setlength{\tabcolsep}{4.2pt}
\resizebox{\textwidth}{!}{
\begin{tabular}{llccccccccccc}
\toprule
\multirow{2}{*}{Method} & \multirow{2}{*}{Input} & \multicolumn{3}{c}{Motion} & \multicolumn{2}{c}{Geometry} & \multicolumn{4}{c}{Language} & \multirow{2}{*}{\makecell{Language\\control}} \\
\cmidrule(lr){3-5} \cmidrule(lr){6-7} \cmidrule(lr){8-11}
 & & Axis Ang $\downarrow$ & Axis Pos $\downarrow$ & Part Motion $\downarrow$ & CD-s $\downarrow$ & CD-m $\downarrow$ & LM-A $\downarrow$ & LSE-A $\uparrow$ & LSP-A $\uparrow$ & State Acc $\uparrow$ & \\
\midrule
Articulate-Anything~\cite{articulate-anything} & monocular video & 25.79 & 4.016 & 32.35 & 158.70 & 440.17 & -- & -- & -- & -- & $\times$ \\
ArtLang~\cite{artlang} & two-state multi-view & \underline{4.85} & \underline{0.103} & \underline{2.32} & \underline{22.05} & \underline{20.98} & \underline{65.97} & \underline{0.897} & \underline{0.701} & -- & \checkmark \\
VideoArtGS~\cite{videoartgs} & monocular video & 15.70 & 1.365 & 7.78 & 33.55 & 151.36 & -- & -- & -- & -- & $\times$ \\
\rowcolor{ourscolor} \ours & monocular video & \textbf{3.30} & \textbf{0.036} & \textbf{0.35} & \textbf{20.11} & \textbf{14.64} & \textbf{51.26} & \textbf{0.904} & \textbf{0.735} & \textbf{0.975} & \checkmark \\
\midrule
\rowcolor{gray!12} Gain & & +1.55 & +0.067 & +1.97 & +1.94 & +6.34 & +14.71 & +0.007 & +0.034 & -- & -- \\
\bottomrule
\end{tabular}}
\end{table*}

\subsection{Persistent Articulation under Motion}
\label{sec:exp_art}

\mA This section scores the joints, the motion they reproduce, the part geometry and the separation of the parts; VideoArtGS and Articulate-Anything have no language interface, so it is the whole comparison for them.

\begin{figure}[t]
\centering
\includegraphics[width=0.8\linewidth]{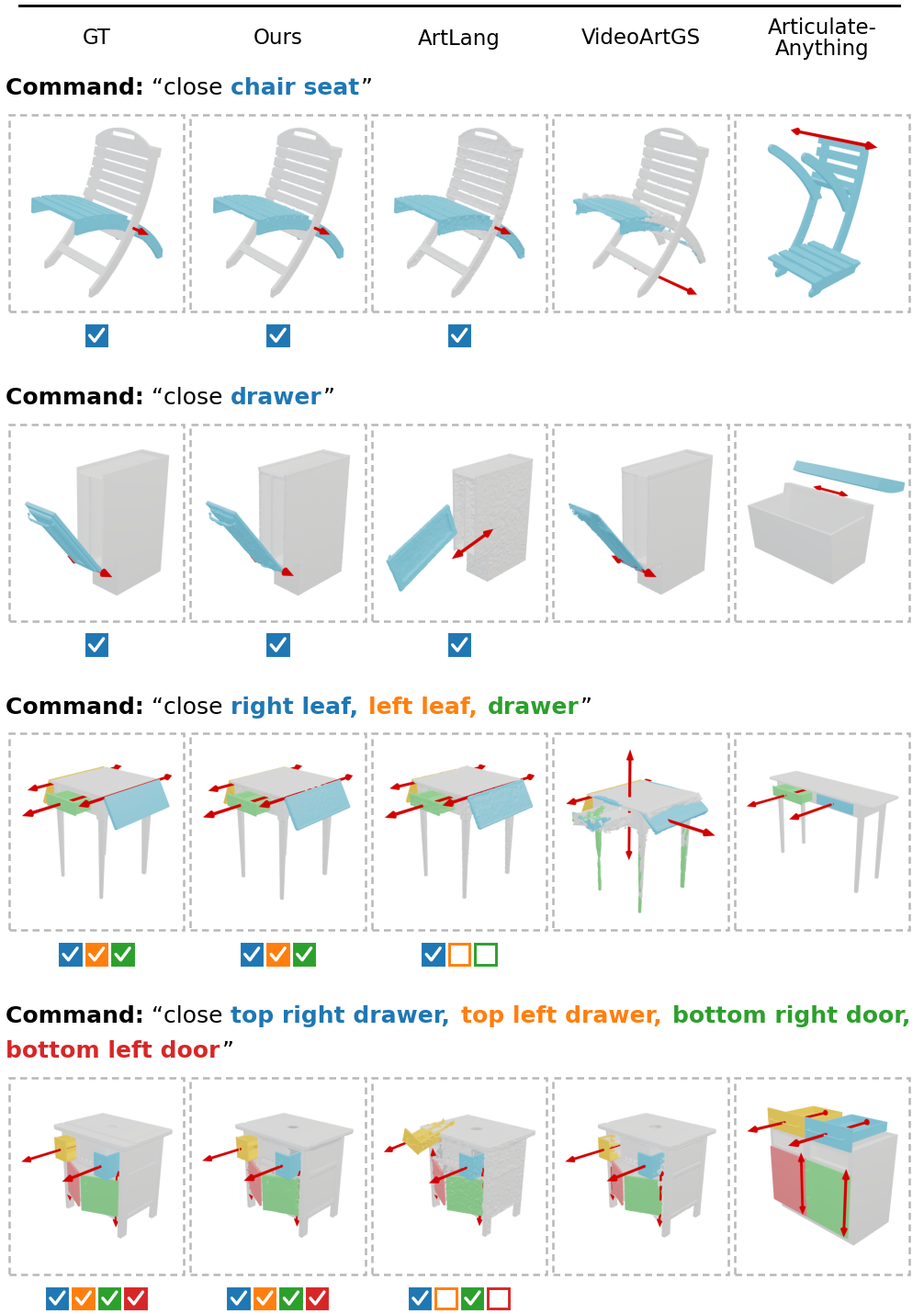}
\caption{\textbf{Articulation and geometry on Articulate-100,} every method at $t=1$ (both end states in Figure~\ref{fig:supp_qual_main} in the appendix). For these objects $t=1$ is the more closed of the two observed states, not mechanically shut. Parts are colored consistently across methods, each predicted part by the true part it is paired with for scoring, arrows show the predicted joint axes, the command names every movable part in its color, and a filled box under a method marks a part that the method gives that name (VideoArtGS does not name parts; Articulate-Anything inherits the template's labels). \ours keeps multi-part objects separated and moves each part along its own axis.}
\label{fig:qual}
\end{figure}

Table~\ref{tab:main} compares all methods. \ours has the lowest error on every motion and geometry metric, although it sees a single video while ArtLang sees both end states from many views. The largest margin is on part motion, $0.35$ against $2.32$ for ArtLang and $7.78$ for VideoArtGS, because the discovered screw joints share one state variable with the video and therefore reproduce the observed amount of motion. VideoArtGS, which shares our input, recovers the static body but often misplaces the movable parts, with a CD-m of $151.36$\,mm. Articulate-Anything observes the same motion, but it predicts joints for a retrieved template rather than for the observed object, and its axis error stays high. Figure~\ref{fig:qual} shows representative objects and the appendix more.

On geometry, the triangle field has the lowest Chamfer distance on both the static and the movable parts, and the movable parts carry the larger gap: a triangle that keeps its part moves rigidly with it, whereas a Gaussian with the wrong part is dragged between two surfaces (Figure~\ref{fig:qual}).

\subsection{State-Dependent Language}
\label{sec:exp_lang}

\begin{figure}[t]
\centering
\includegraphics[width=0.66\linewidth]{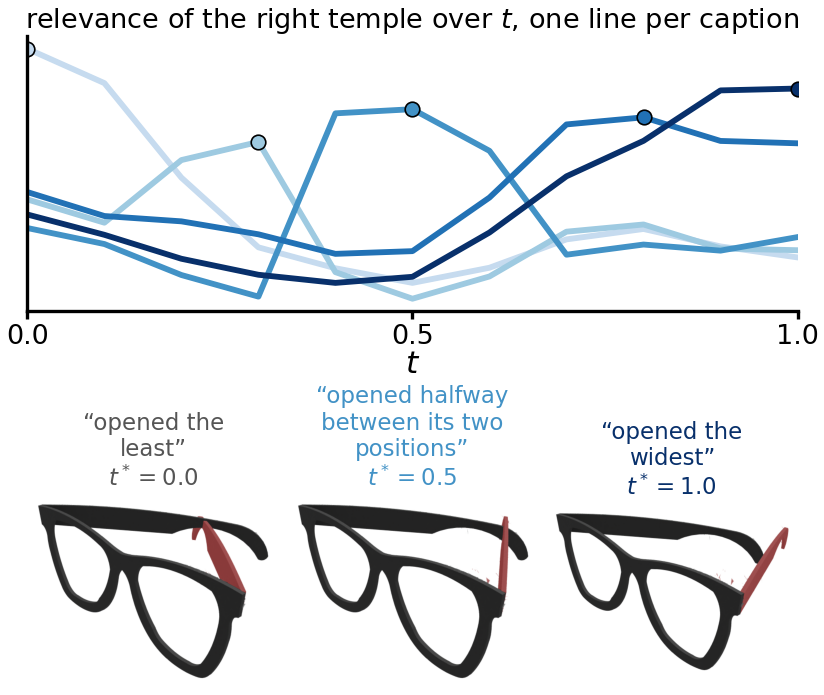}
\caption{\textbf{Queries that differ only in the state.} All three select the same part; the state at which its relevance peaks follows the state word, and the command poses the part there.}
\label{fig:queries}
\end{figure}

\mL\ours is the only method that answers state queries, and its per-part state classification, State Acc, is $97.5\%$ (Table~\ref{tab:main}). Figure~\ref{fig:queries} shows queries that differ only in their state words. The relevance curves of the queried part peak at different states, and the executed commands pose the part accordingly. Asked every caption as a free query over all parts and states, \ours answers $64.2\%$ correctly against $19.4\%$ for the static field (Sec.~\ref{sec:supp_results}, Figure~\ref{fig:supp_queries}).

\subsection{Executable Temporal Grounding}
\label{sec:exp_ctrl}

\mT LM-A scores naming and motion together, since it moves the named part along the recovered joint; LSE-A and LSP-A score the naming alone. \ours also improves the name-based metrics of ArtLang, with $0.904$ against $0.897$ LSE-A, $0.735$ against $0.701$ LSP-A, and an LM-A of $51.26$ against $65.97$\,mm.

\begin{table}[t]
\centering
\caption{\textbf{Ablations on Articulate-100.} \emph{Static language} keeps the full model's geometry and joints (--) and drops the state captions and the $t$-conditioned residual; \emph{free deformation} replaces the screw joints by a deformation field over $(x,t)$; the naming rows change only how names are assigned.}
\label{tab:ablation}
\setlength{\tabcolsep}{2.6pt}
\resizebox{\linewidth}{!}{
\begin{tabular}{lccccccccc}
\toprule
Variant & \makecell{Axis\\Ang $\downarrow$} & \makecell{Axis\\Pos $\downarrow$} & \makecell{Part\\Motion $\downarrow$} & \makecell{CD-s\\$\downarrow$} & \makecell{CD-m\\$\downarrow$} & \makecell{LM-A\\$\downarrow$} & \makecell{LSE-A\\$\uparrow$} & \makecell{LSP-A\\$\uparrow$} & \makecell{State\\Acc $\uparrow$} \\
\midrule
\rowcolor{ourscolor} Full & \textbf{3.30} & \textbf{0.036} & \textbf{0.35} & \textbf{20.11} & \textbf{14.64} & \underline{51.26} & \underline{0.904} & \textbf{0.735} & \textbf{0.975} \\
Static language & -- & -- & -- & -- & -- & \textbf{50.95} & \textbf{0.906} & \textbf{0.735} & \underline{0.141} \\
Free deformation & \underline{25.79} & \underline{0.613} & \underline{9.74} & \underline{22.50} & \underline{343.13} & 234.82 & 0.810 & 0.645 & \textbf{0.975} \\
w/o type-aware naming & -- & -- & -- & -- & -- & 52.49 & 0.903 & \underline{0.733} & -- \\
Position-only naming & -- & -- & -- & -- & -- & 53.55 & 0.896 & 0.722 & -- \\
\bottomrule
\end{tabular}}
\end{table}

\subsection{Ablations}
\label{sec:exp_abl}

\begin{figure}[t]
\centering
\includegraphics[width=0.62\linewidth]{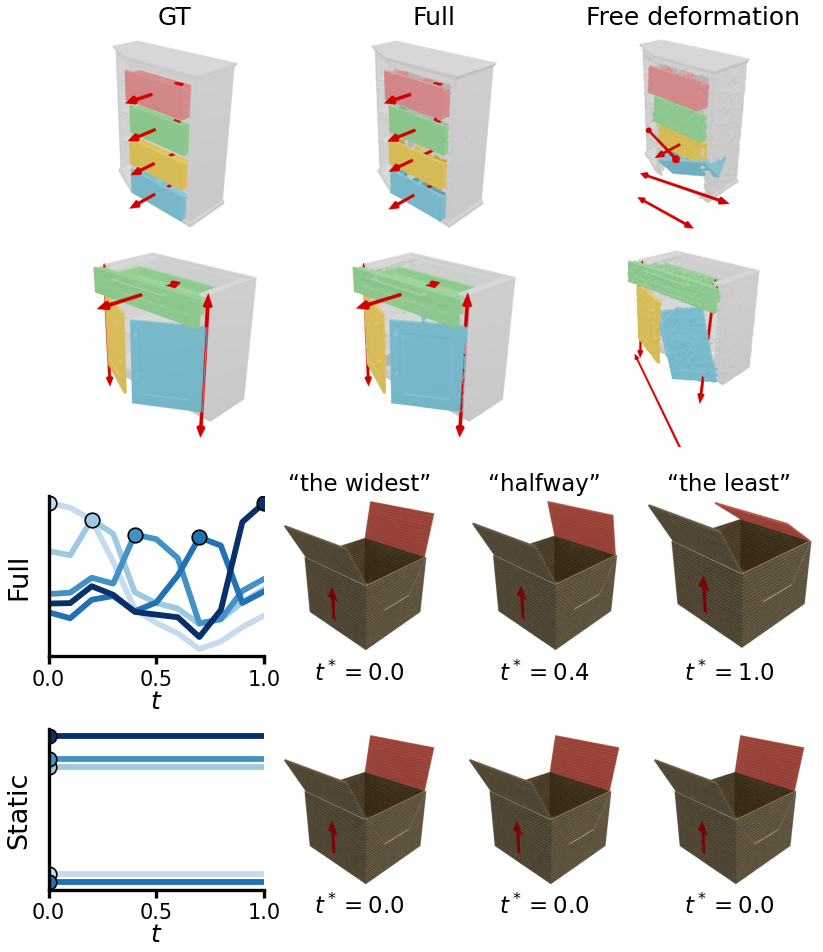}
\caption{\textbf{Ablations.} A free deformation fits the frames but breaks the parts between them (top); without state conditioning the part is right but its relevance is flat over $t$ (bottom).}
\label{fig:ablation}
\end{figure}

Table~\ref{tab:ablation} and Figure~\ref{fig:ablation} isolate the design choices.

\mA\textbf{Screw joints versus free deformation.} Replacing the joints by a deformation field over $(x,t)$, as in 4D language fields~\cite{4dlangsplat}, raises the axis error from $3.30^\circ$ to $25.79^\circ$ and CD-m from $14.64$ to $343.13$\,mm. The deformation fits the frames without keeping each part rigid, and joints fit to it afterwards inherit its errors, raising LM-A to $234.82$\,mm. State Acc is unchanged because the language stage sees the same captions and frames, which shows that state retrieval alone does not test whether the motion is right.

\mL\textbf{State conditioning.} Removing the state captions and the residual of Eq.~\eqref{eq:lang} leaves the name-based metrics unchanged within noise, since the variant shares the geometry and joints and names do not depend on the state. State Acc, however, falls from $0.975$ to $0.141$. In a static field the feature of a part is the same at every state, so in all frames of that part it matches the same one of the five captions, and the state cannot be read from it. This is what the $t$-conditioned residual fixes.

\mT\textbf{Type-aware naming.} Dropping the joint-type term of Eq.~\eqref{eq:naming} raises LM-A from $51.26$ to $52.49$\,mm, and assigning names by projected image position alone, the rule of ArtLang, raises it further to $53.55$\,mm. The type term separates neighboring parts that look alike but move differently, such as a door above a drawer; parts that look and move alike remain the main failure (Sec.~\ref{sec:supp_failures}).

\begin{figure}[t]
\centering
\includegraphics[width=0.78\linewidth]{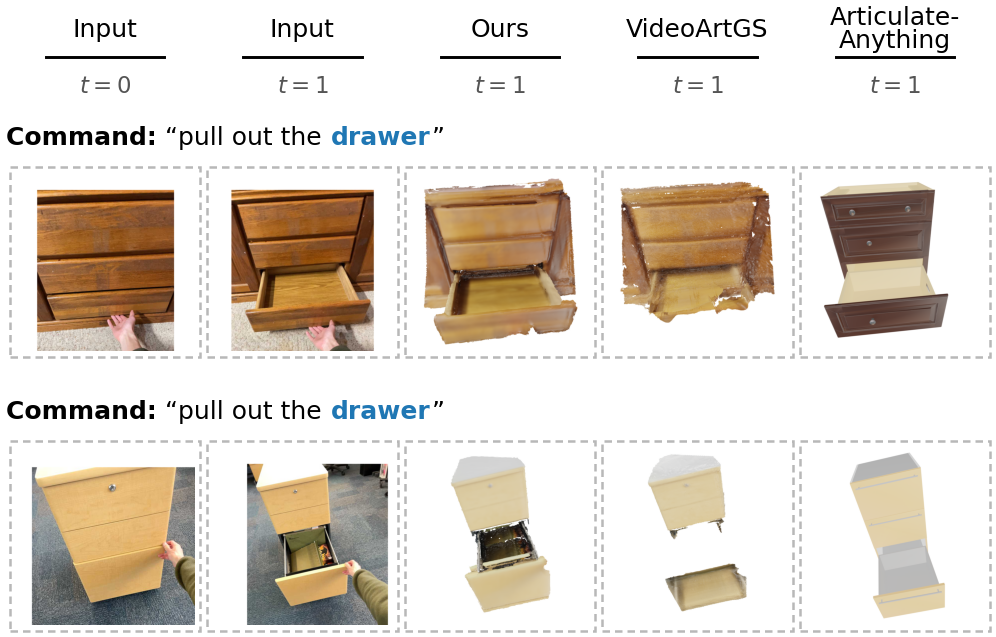}
\caption{\textbf{Real captures.} The first and last frame of the interaction, then ours, VideoArtGS and Articulate-Anything at $t=1$. Ours follows the input; VideoArtGS floats in front of the measured depth; Articulate-Anything shows a retrieved template.}
\label{fig:real}
\end{figure}

\subsection{Real Captures}
\label{sec:exp_real}

We also apply \ours to the real phone captures released with iTACO~\cite{video2articulation}, which have no ground truth (Figure~\ref{fig:real}). A hand pulls a drawer out while the camera moves; poses and metric depth are iTACO's, SAM~3~\cite{sam3} segments the hand and the parts, and the moved part receives a learned articulation time. On both objects \ours finds one prismatic part along the drawer's travel and \emph{pull out the drawer} moves the right part. ArtLang needs both end states from many views and is left out (Sec.~\ref{sec:supp_real} in the appendix).

\section{Conclusion}
\label{sec:conclusion}

\ours reconstructs an articulated object from one monocular RGB-D video and attaches a language field indexed by articulation time, a joint state, rather than video time. Screw joints discovered from 3D part observations give the state a kinematic meaning, comparative captions tie it to language, and a query returns a part and a state that can be executed on the asset while the other parts keep theirs. On Articulate-100 this yields more accurate joints and geometry than reconstruction and articulation baselines and a state retrieval accuracy a state-agnostic field cannot reach, and the same pipeline runs on phone captures. The method assumes one degree of freedom per part and two observed end states; longer interaction videos are a natural next step.

\clearpage
{
    \small
    \bibliographystyle{ieeenat_fullname}
    \bibliography{main}
}

\clearpage
\appendix
\renewcommand{\dblfloatpagefraction}{0.5}
\setcounter{dbltopnumber}{3}
\section{Implementation Details}
\label{sec:supp_impl}

\textbf{Canonical triangle field.} The triangle field starts with one triangle per point of a cloud fused from the depth of the input video. Each triangle is an isosceles right triangle inscribed in a circle whose radius is $2.23$ times the distance to the nearest neighboring point, with a random orientation. Colors use spherical harmonics up to degree three, raised by one degree every 1{,}000 iterations, and vertex opacities start at $0.28$. The window sharpness $\sigma$ of the rasterizer is annealed from $1.0$ to $0.3$ over the whole schedule. Every 500 iterations, until $60\%$ of the schedule, triangles whose opacity or blending weight falls below a threshold are pruned and the triangles with the highest blending importance are subdivided, up to $500$k vertices. An opacity floor rises from $0.1$ to $0.9999$ between iteration 5{,}000 and $80\%$ of the schedule, which makes the field opaque. The rasterizer runs at $2\times$ resolution from $2/3$ of the schedule and at $4\times$ two thousand iterations later, and the image is downsampled by area averaging. After training, every triangle whose largest blending weight over all input frames stays below $0.5$ is removed.

\textbf{Part identity field.} The planes $\varphi$ in Eq.~\eqref{eq:weights} are the $xy$, $xz$ and $yz$ planes of a HexPlane at four resolutions ($64$, $128$, $256$ and $512$ cells, $32$ channels each). Their features are multiplied across planes and concatenated across resolutions into a 128-dimensional vector, which a linear layer and a two-layer MLP of width 64 map to $K+1$ logits. Before the articulated stage, the decoder is fitted for 500 steps to the part labels of the canonical points $X_k$.

\textbf{Schedule and losses.} Table~\ref{tab:supp_hparams} lists the iteration counts, learning rates and loss weights. The coarse stage uses only the rest-state frames ($t_i=0$) and no motion. The articulated stage admits frames with $t\le\min(1,\,2\tau)$ at normalized step $\tau$, so the full range is available after half of the stage. $\mathcal L_{\rm occ}$ is $-\log(1-\hat\alpha)$ on background pixels whose color matches the background, $\mathcal L_{\rm reg}$ holds an opacity sparsity term ($1.9\times10^{-6}$, first 5{,}000 iterations), an opacity entropy term ($0.01$), a normal consistency term ($5\times10^{-5}$, after 5{,}000 iterations), an entropy term on the part weights ($10^{-3}$), and the second-order smoothness of the HexPlane planes ($10^{-4}$).

\begin{table}[t]
\centering
\caption{\textbf{Hyperparameters.} The same values are used for every object.}
\label{tab:supp_hparams}
\setlength{\tabcolsep}{4pt}
\resizebox{\linewidth}{!}{
\begin{tabular}{ll}
\toprule
Item & Value \\
\midrule
Joint discovery & 6 starts, 3{,}000 free + 2{,}000 locked steps, Adam $10^{-2}$ \\
Type refit & 600 steps per hypothesis, margin $m=0.1$ \\
Cauchy scale $\sigma$ & $0.07\,\ell \rightarrow 0.0035\,\ell$, geometric \\
Coarse / articulated stage & 10{,}000 / 14{,}000 iterations \\
Language stage & 5{,}000 iterations \\
Part label refinement & 1{,}500 iterations, Adam $0.05$ \\
Vertex learning rate & $1.5\times10^{-3}\rightarrow1.5\times10^{-5}$ \\
Color / opacity learning rate & $1.6\times10^{-3}$ / $3\times10^{-2}$ \\
Joint / part decoder learning rate & $10^{-3}$ / $10^{-3}$ \\
Language feature / $\delta$ learning rate & $2.5\times10^{-3}$ / $1.6\times10^{-4}\times$ camera extent \\
$\lambda_{\rm SSIM}$, $\lambda_d$, $\lambda_o$, $\lambda_p$, $\lambda_c$ & $0.2$, $0.2$, $0.5$, $0.05$, $1$ \\
Joint guard & revert if $E_k^{\rm trained}>1.25\,E_k^{\rm disc}$ \\
Caption buckets $B$, language dimension $d$ & 5, 24 \\
Query grid $\mathcal G$ & 11 states on $[0,1]$ \\
\bottomrule
\end{tabular}}
\end{table}

\textbf{Joint transform field.} For each start, $(\boldsymbol\omega_k,\mathbf v_k,\mathbf p_k)$ are optimized with both indicators on by Eq.~\eqref{eq:corr} on one random frame per step. Start one places the pivot at the centroid of $X_k$ with zero rotation and translation. The other five use a rotation of $0.3$\,rad and a translation of $0.05\,\ell$ along random directions and a pivot drawn uniformly in the part's bounding box. The revolute hypotheses of Eq.~\eqref{eq:type} are the screw axis of the fitted motion at $t=1$, moved to the point of the axis nearest the part centroid, and a zero rotation about that centroid. The prismatic hypothesis slides along the displacement of the centroid. The score $E_k$ used by the multi-start selection and by the joint guard is the mean of part $k$'s term in Eq.~\eqref{eq:corr} over all frames with $t>0$, at $\sigma=0.0035\,\ell$ and with a fixed point subsample.

\textbf{3D part observations.} A part region is eroded by two pixels, pixels whose depth gradient exceeds $3\%$ of the depth range are removed, and the rest are back-projected. A part needs at least 16 pixels in a frame. Each observation is subsampled to 2{,}048 points and each canonical set $X_k$ to 8{,}192.

\textbf{State-dependent language field.} The per-triangle features start as $0.01\,\mathcal N(0,I)$. $\delta$ is an MLP with two hidden layers of width 64. A frame is sampled at random in every iteration, and its target is the bank entry of the bucket nearest its state. Geometry, part decoder and joints receive no gradient.

\textbf{Compute.} One object takes about two hours on a single NVIDIA A800 GPU, including discovery, the two geometry stages and the language stage.

\section{Caption Bank}
\label{sec:supp_captions}

\textbf{Part names and opening direction.} GPT-4o receives two renders of the object from one camera, at $t=0$ and $t=1$, together with the request to list the movable parts with their joint types and to state which image shows the parts opened wider. The question is asked a second time with the two images swapped. A consistent answer is kept; otherwise the image in which the parts cover a larger area decides. The names are matched to the mask labels by a Hungarian assignment between image centroids and the boxes GPT-4o returns.

\textbf{Phrase scales.} Table~\ref{tab:supp_phrases} lists the five-step scales. The scale is chosen by the joint type and, for revolute joints, by the name (knobs, wheels and dials turn). Bucket $b$ of a part receives phrase $\min(4,\lfloor5\eta_b\rfloor)$ of its scale. If two parts receive the same caption, a location suffix (\emph{on the left side}, \emph{in the center}, \emph{on the right side}) taken from the horizontal mask centroid separates them.

\begin{table}[t]
\centering
\caption{\textbf{Comparative state phrases} by motion, from the least to the most open state.}
\label{tab:supp_phrases}
\small
\setlength{\tabcolsep}{4pt}
\begin{tabular}{l p{0.78\linewidth}}
\toprule
Motion & Phrases, $\eta_b$ from low to high \\
\midrule
swings & opened the least; opened a little; opened halfway between its two positions; opened wider; opened the widest \\
\addlinespace[2pt]
slides & pulled out the least; pulled out a little; pulled out halfway between its two positions; pulled out further; pulled out the furthest \\
\addlinespace[2pt]
turns & turned the least; turned a little; turned halfway between its two positions; turned further; turned the furthest \\
\bottomrule
\end{tabular}
\end{table}

\textbf{Whitened PCA.} The basis of Eq.~\eqref{eq:embed} is fitted per object on all its captions, one name phrase per part, and a fixed vocabulary of 195 phrases (20 part nouns, 15 state phrases, the first eight nouns combined with every state phrase, four phrases on visibility and motion, and nine location words combined with four nouns), so that it spans the words a user is likely to use and not only the object's own captions. Whitening divides every principal direction by its singular value. Without it, the few directions that dominate short part phrases leave \emph{left door} and \emph{right door} almost identical after projection.

\section{Metrics and Protocol}
\label{sec:supp_metrics}

\textbf{Motion.} Axis Ang is the angle between the predicted and true axis directions, ignoring sign. Axis Pos is the distance between the predicted and true axis lines for revolute joints, in units of $0.1$\,m; a prismatic joint has no axis line and contributes zero to the per-object mean, the convention of the released evaluation code of GaussianArt~\cite{gaussianart} and ArtGS~\cite{artgs}. Part Motion is the geodesic rotation error at $t=1$ in degrees for revolute joints and the translation error in meters for prismatic joints. For every method, the predicted movable parts are paired one-to-one with the true movable parts by a Hungarian assignment on the Chamfer distance between them, averaged over the two end states, which is the pairing CD-m uses, and the joint metrics inherit that pairing; the static parts are compared directly, and a true part left without a partner scores the per-part penalty. Errors are averaged over the parts of an object and then over objects. A predicted joint is scored under its own type, not the true one. A method that declares its joint types keeps them, and a method that outputs a rigid transform per part is read as a hinge when the transform rotates by $5^\circ$ or more and as a slide otherwise, so that a slide whose fitted transform carries a small rotation is not scored as a hinge. A part whose type differs from the true joint's scores the angle between its own axis, the hinge axis of a predicted hinge or the slide direction of a predicted slide, and the true axis, the axis-position penalty when the true joint is a hinge, and the part-motion penalty of a failed part.

\textbf{Geometry.} CD-s and CD-m are symmetric Chamfer distances in millimeters between 10{,}000 points sampled on the predicted and true static and movable parts, averaged over the two end states. All reconstructed geometry, ours included, is scored on surfaces extracted by TSDF fusion. For our method and ArtLang we fuse the depth of each part, rendered in both end states from 96 views spread evenly on a sphere ($640\times640$ pixels, $40^\circ$ field of view), into a volume in the part's rest frame with 320 voxels across the object's bounding-box diagonal and a truncation of three voxels, extract its surface, remove components smaller than $2\%$ of the largest, apply eight Taubin smoothing iterations and decimate it to at most 150k triangles, and VideoArtGS uses the TSDF meshes of its own release. Articulate-Anything retrieves its parts as meshes from a template library rather than reconstructing them, so its template part meshes are scored directly.

\textbf{Language.} LSE-A and LSP-A follow ArtLang~\cite{artlang}. The named parts are rendered at the held-out cameras of Articulate-100 in both end states. Every name is split into a noun key and a spatial key, and for each key the IoU is taken between the union of predicted parts carrying it and the union of true parts carrying it. LSP-A is defined only for objects whose true names contain a spatial word. LM-A pairs predicted and true parts by name agreement, moves both by $\alpha\in\{-1,-\tfrac12,0,\tfrac12,1\}$ of their joint, and averages the Chamfer distance over $\alpha$ and joints. State Acc is described in Sec.~\ref{sec:setup}.

\textbf{Failures.} Following iTACO~\cite{video2articulation}, a method that fails on an object receives $90^\circ$ for axis angle and revolute part motion, $1$\,m for prismatic part motion and axis position, and $1{,}000$\,mm for the Chamfer distances. Extending this rule per part, a true part that a method does not predict receives the same penalty on that part only.

\section{Baselines}
\label{sec:supp_baselines}

\textbf{VideoArtGS}~\cite{videoartgs} receives the same 172 frames, depth in millimeters, cameras converted to its convention, and the state $t$ as its per-frame time variable. The 100 rest-state frames form its canonical set and the remaining frames its articulation set. Its joint count and types come from our GPT-4o step instead of its own query, and its initial cloud is back-projected from the rest frames, as in its own pipeline. We run its released code unchanged: 3D tracks, 20{,}000 canonical iterations, 10{,}000 deformation initialization iterations and 20{,}000 joint training iterations. Its parts are paired with the true parts by the Chamfer assignment of Sec.~\ref{sec:supp_metrics}, as for every method.

\textbf{ArtLang}~\cite{artlang} uses its released models, trained with its ArtMesh backend on posed multi-view RGB-D images of the two end states of each object with the rendered part masks, the same supervision our method receives from one video.

\textbf{Articulate-Anything}~\cite{articulate-anything} receives the stretch of each video in which the parts move, as RGB frames without depth or cameras. We run its released code with GPT-4o in place of its default Gemini 1.5 Flash; retrieval is its own CLIP category filter followed by the vision-language model. The test object is removed from its retrieval library, so it retrieves a different PartNet-Mobility template and predicts the joints of that template. Its video mode predicts by default only the joint of the part it sees moving, which we keep for objects with one movable part; on objects with several movable parts we switch to its all-joints mode so that every joint is predicted in one run. Its predicted joints are taken in the global frame of its output and rotated back by the $180^\circ$ turn about the vertical axis that its preprocessing applies. Without depth the template has no metric scale or position, so before scoring we center it on the bounding box of the true object at rest and scale it to the same diagonal. Its parts are then paired with the true parts by the same Chamfer assignment. Its two states are the template's rest pose and the limit of the predicted joint, in that order whichever way the video went, so it cannot tie them to the observed states. We therefore score each object under both orderings of its states, its own and the reversed one, and keep the one with the lower part-motion error; 66 of the 100 objects take the reversed ordering, and its renders in Figure~\ref{fig:qual} and in Figures~\ref{fig:supp_qual_main}, \ref{fig:supp_qual}, \ref{fig:supp_qual_2} and~\ref{fig:supp_fail} show the same ordering. Axis angle and axis position do not depend on the ordering.

\section{Additional Results}
\label{sec:supp_results}

\textbf{Query-level retrieval.} State Acc scores one part at a time against its five captions. A stricter test asks every caption as a free query and solves Eq.~\eqref{eq:query_grounding} over all parts and the 11-state grid, counting it correct only when both the part and the bucket of $t^*$ are right. Averaged over the 100 objects, \ours answers $64.2\%$ of these queries correctly, while the static-language variant answers $19.4\%$, as expected from Sec.~\ref{sec:exp_abl}. Its relevance curves are exactly flat in $t$, so it returns the same state for all five captions of a part. Accuracy is highest on storage furniture and lowest on tables, whose drawers and legs often look alike (Sec.~\ref{sec:supp_failures}).

\begin{figure*}[t]
\centering
\includegraphics[width=\textwidth]{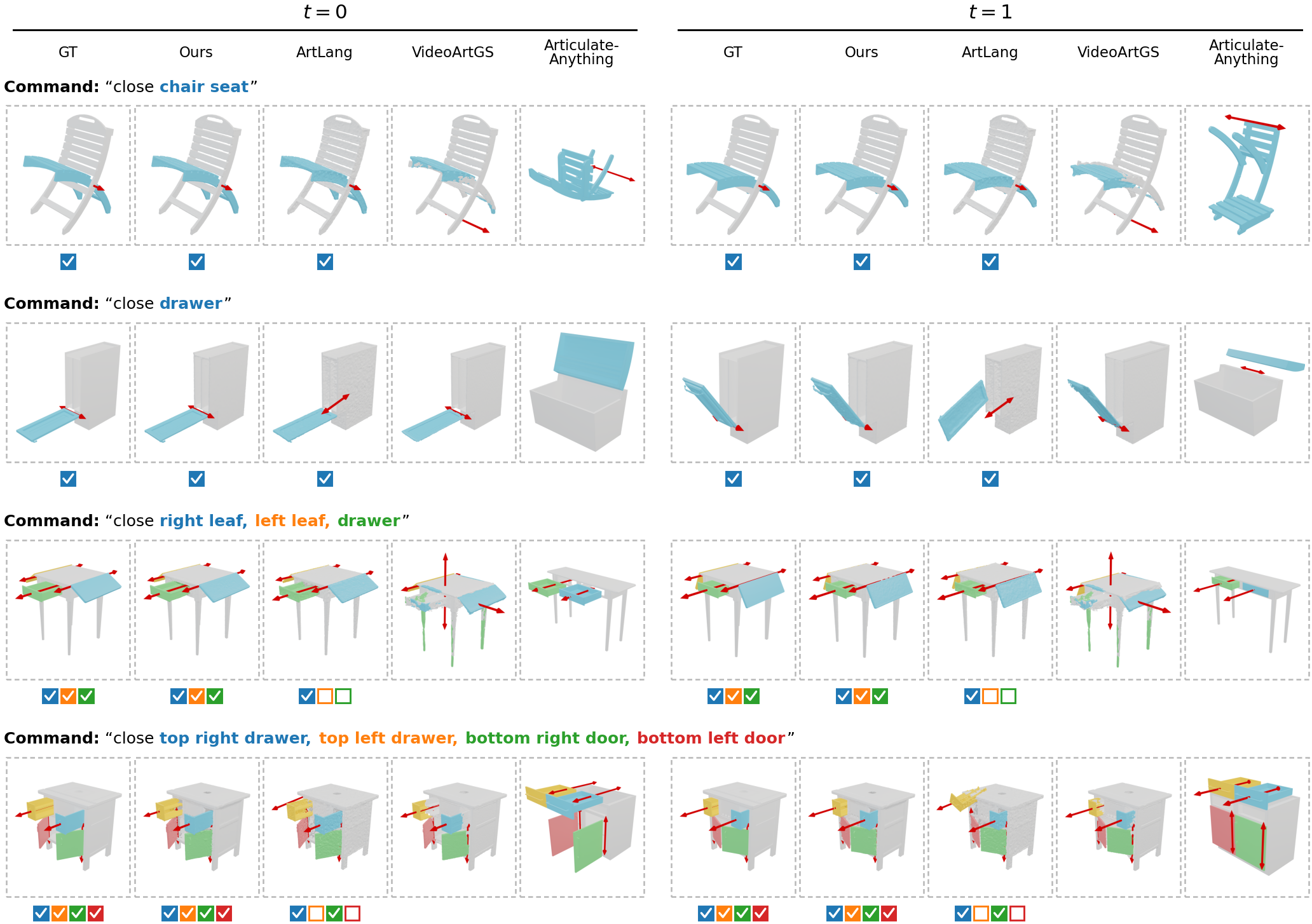}
\caption{\textbf{The objects of Figure~\ref{fig:qual} at both end states,} every method at $t=0$ (left) and $t=1$ (right), with the same colors, axes and name boxes.}
\label{fig:supp_qual_main}
\end{figure*}

\begin{figure*}[t]
\centering
\includegraphics[width=0.9\textwidth]{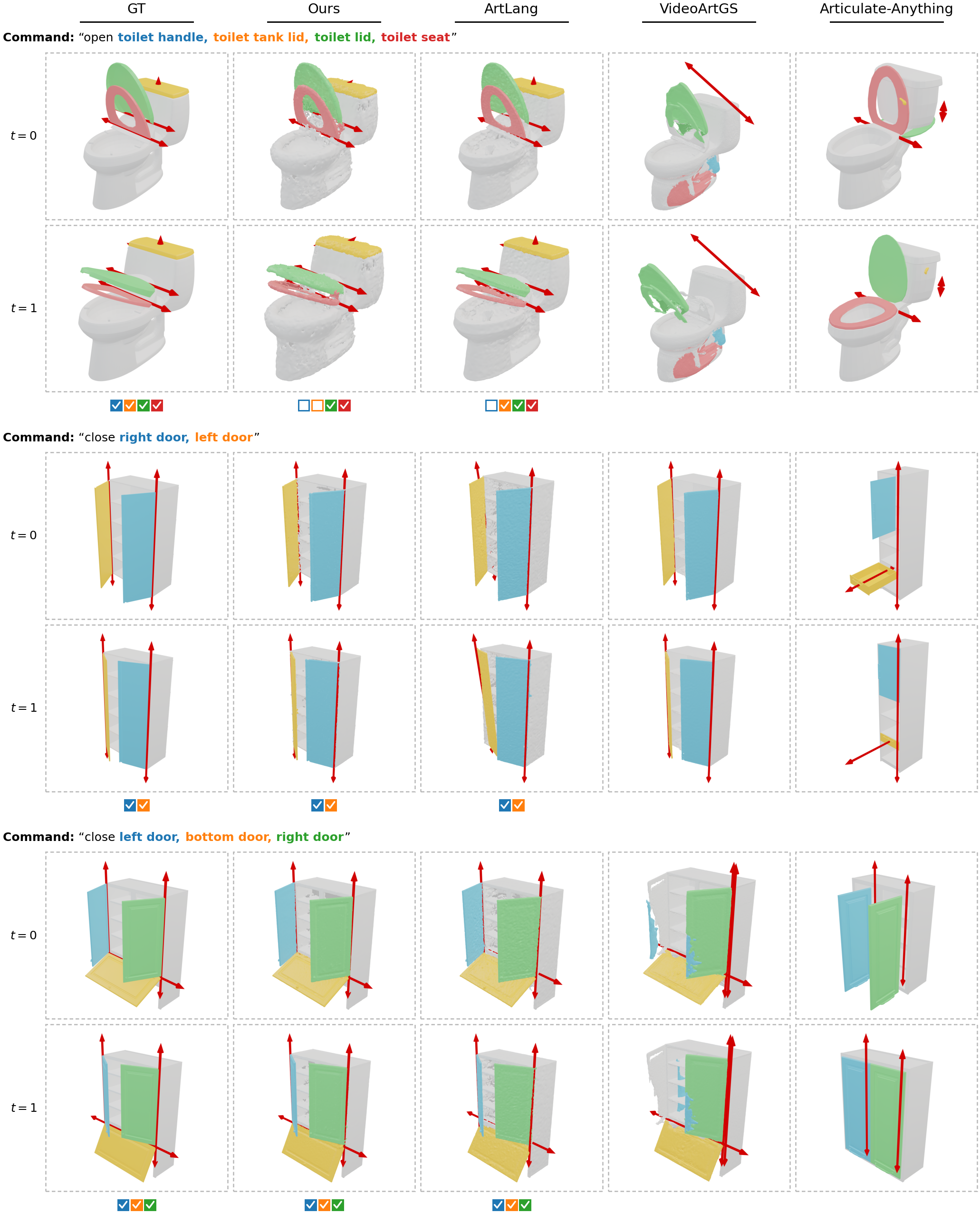}
\caption{\textbf{More results on Articulate-100 (1/2).} Both end states, joint axes as arrows, parts colored by the ground-truth palette, each predicted part by the true part it is paired with for scoring. A filled box under a method marks a part that the method gives the name it has in the command.}
\label{fig:supp_qual}
\end{figure*}

\begin{figure*}[t]
\centering
\includegraphics[width=0.9\textwidth]{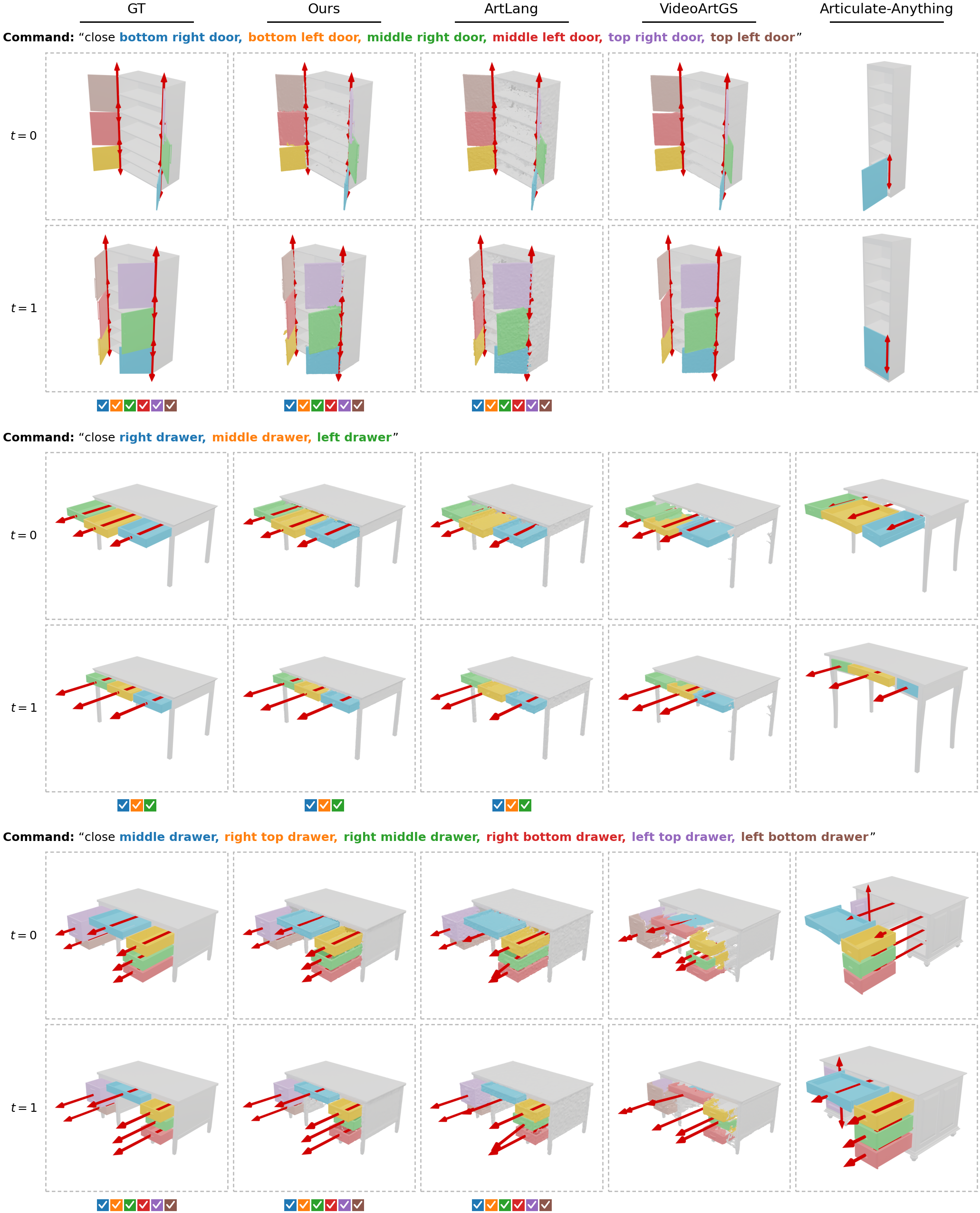}
\caption{\textbf{More results on Articulate-100 (2/2).} Both end states, joint axes as arrows, parts colored by the ground-truth palette, each predicted part by the true part it is paired with for scoring. A filled box under a method marks a part that the method gives the name it has in the command.}
\label{fig:supp_qual_2}
\end{figure*}

\textbf{More qualitative comparisons.} Figure~\ref{fig:supp_qual_main} shows the objects of Figure~\ref{fig:qual} at both end states, and Figures~\ref{fig:supp_qual} and~\ref{fig:supp_qual_2} extend it to further objects.

\begin{figure*}[t]
\centering
\includegraphics[width=0.85\textwidth]{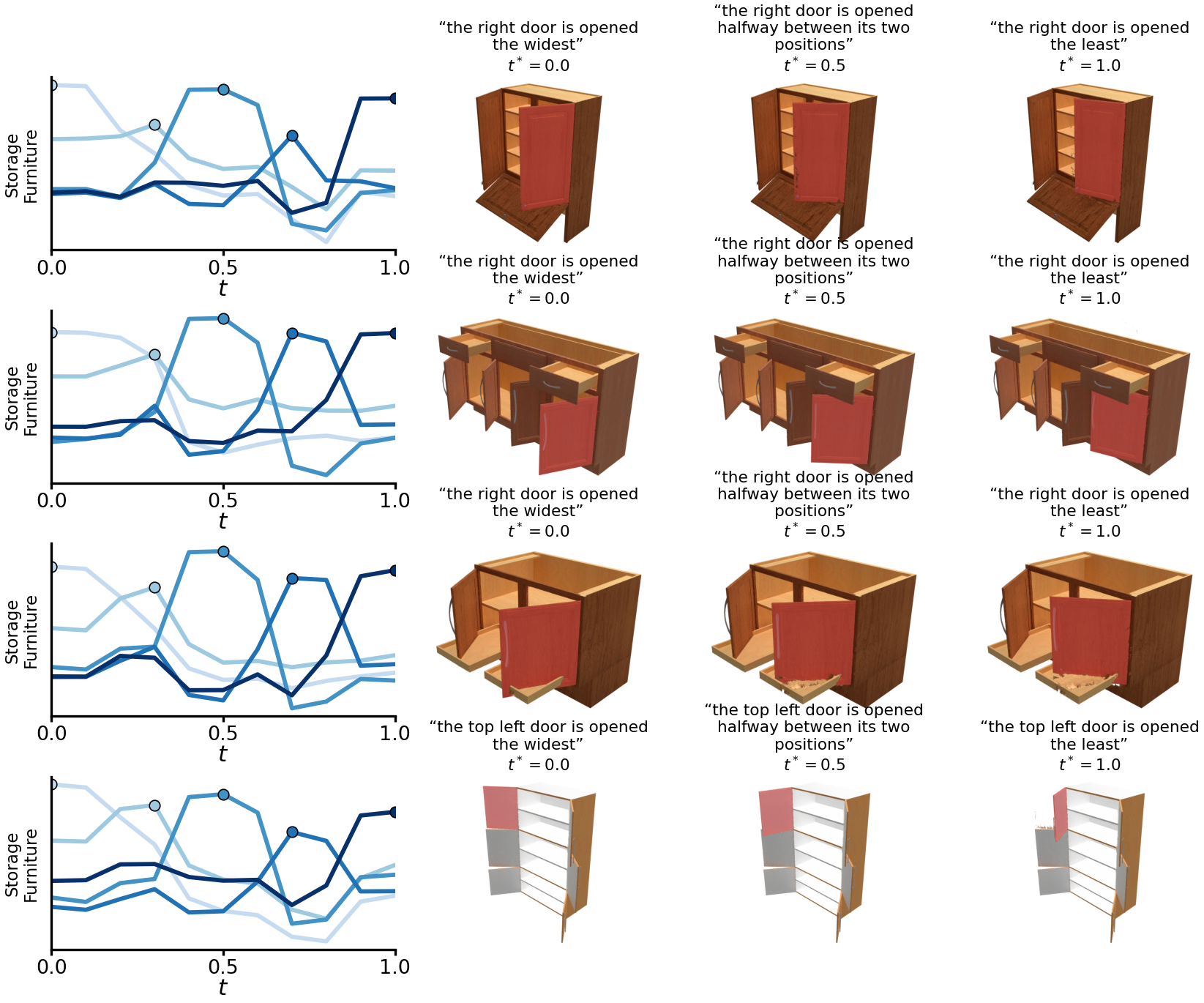}
\caption{\textbf{State queries on further objects.} Left: relevance of the queried door over $t$ for the five captions of its scale, each peaking near the state it describes. Right: the states retrieved for three of the captions.}
\label{fig:supp_queries}
\end{figure*}

\textbf{More state queries.} Figure~\ref{fig:supp_queries} shows state queries on four further storage objects with several doors, where the query has to single out one of them.

\section{Real-Capture Pipeline}
\label{sec:supp_real}

The captures are the real RGB-D videos of iTACO~\cite{video2articulation}, handheld iPhone videos recorded with Record3D, with the camera poses exported by the recording app and the per-frame metric depth of iTACO's preprocessing, which upsamples the phone's low-resolution LiDAR depth with Prompt Depth Anything. SAM~3~\cite{sam3} segments the hand, the object, and every instance of a part noun such as \emph{drawer}. Hand pixels are marked as unknown and excluded from every loss, so the hand neither carves holes into the parts it touches nor paints them in its color.

\textbf{Parts from hand contact.} The part a person moves is the instance that the dilated hand overlaps most. Consecutive frames of contact with the same instance form a run. A run in which the instance moves by at least 2\,cm and that lasts at least 8 frames becomes a part, numbered in the order the parts are moved. A run that moves nothing, such as a hand steadying the object, is ignored. A later run continues an earlier part only when its points and that part's final points match in both directions and their centers lie within 3\,cm, so that neighboring drawers pushed one after the other stay separate parts. Each part is then labeled in every frame by matching the 3D points of each instance to the part's points at that frame's state, before, during or after its own motion.

\textbf{Per-part articulation time.} A real frame has no known state. Each part starts from $t_k=0$ before its run, the normalized displacement of its points during the run, and $1$ after it; displacement below 1\,cm counts as sensor noise, so the frames before the hand moves a part keep $t_k=0$. Training learns a per-frame offset for every part, pulled towards this estimate by an $L_2$ term of weight $0.1$ and optimized with Adam at a learning rate of $2\times10^{-3}$ from $10\%$ of the articulated stage on, so a part moved later can stay at rest while another one moves.

\textbf{Display of the real reconstructions.} The capture's depth is noisy and the camera never sees the object from all sides, so the triangle field fitted to it carries stray triangles that make a figure hard to read. For Figure~\ref{fig:real} we therefore show the captured depth itself, organized by what the model learned: every frame's depth, restricted to a part's label and moved back to the rest state with the inverse of the learned joint motion at that frame's articulation time, is integrated into one truncated signed distance volume per part; the body takes every frame, with the pixels of the moving parts and of the hand excluded. Pixels at depth discontinuities and within six pixels of the background, the hand or another part are dropped, small disconnected pieces are removed, and the surfaces are lightly smoothed. The parts, the joints and the articulation times are those of the learned model, only the surface comes from the depth, so a large error in a joint or a label still appears as a doubled or displaced part, although the filtering and smoothing can hide small ones. The figure therefore shows the recovered parts, joints and articulation on real data, not the quality of the learned surface, which is measured on the synthetic objects. Articulate-Anything predicts joints for a retrieved template in the template's own frame, so for the figure we place the template in the capture's frame with one similarity transform fitted to the observed points. The joints are its own. ArtLang is not shown on the real captures, since it needs posed multi-view RGB-D images of both end states, which one handheld video does not provide.

\section{Failure Cases}
\label{sec:supp_failures}

\begin{figure*}[t]
\centering
\includegraphics[width=\textwidth]{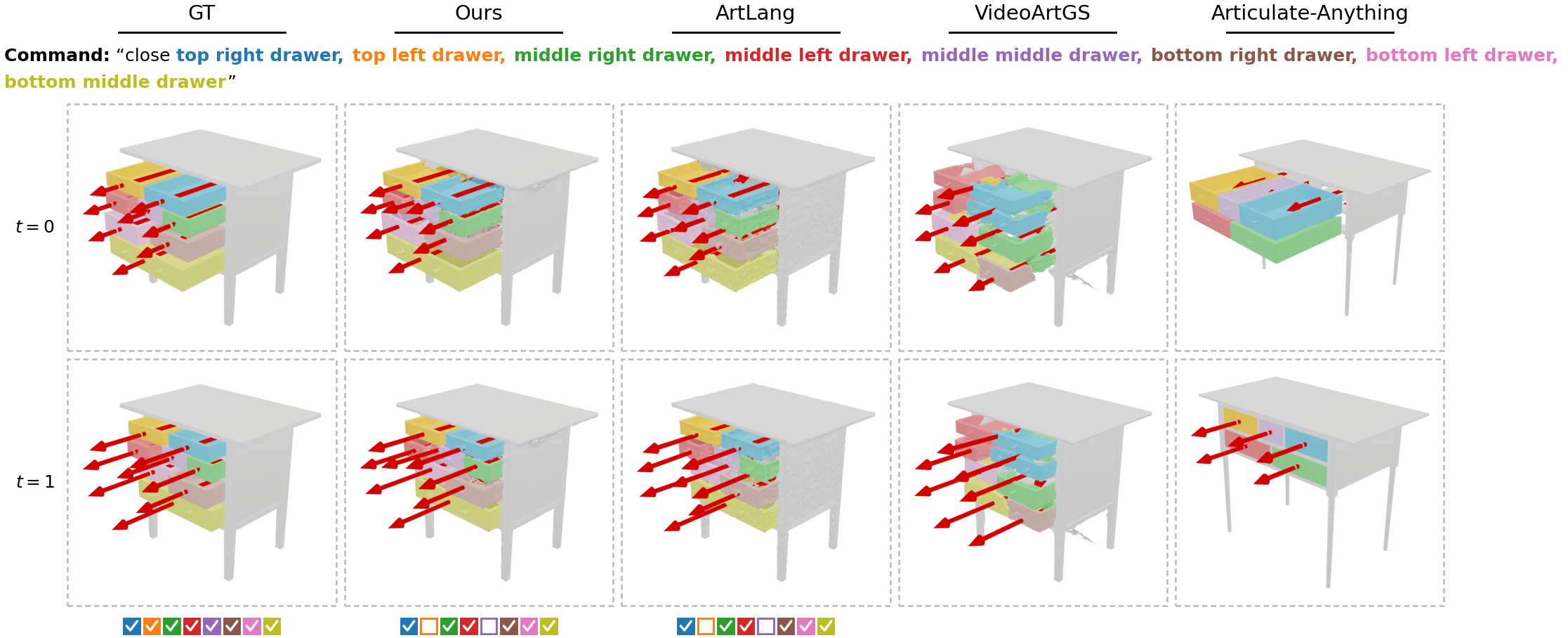}
\caption{\textbf{Failure case: repeated similar parts.} A table with eight drawers in three rows, in both end states. A filled box under a method marks a drawer that the method gives the name it has in the command, and an empty box a drawer whose name is assigned to a neighbor.}
\label{fig:supp_fail}
\end{figure*}

\textbf{Repeated similar parts.} Our naming can fail on objects with many parts that look alike and differ only by position. Figure~\ref{fig:supp_fail} shows a table with eight drawers in three rows. Their names differ only by location words, and the part each drawer receives is decided by its position and by how its language feature compares with the captions of its neighbors. Our naming gives six of the eight drawers their own names, while the top left and middle middle drawers receive other names, and ArtLang misses the same two. The joints are unaffected, so the drawers still move correctly, but a command that names one drawer can move a neighboring one.

\section{Additional Baselines}
\label{sec:supp_more_baselines}

\textbf{FAMOS.} FAMOS~\cite{famos} is a feed-forward model that predicts the movable parts of an object and one joint per part from a few partial point clouds of the object in different states. We give it the observations our method receives, the depth of the video back-projected with its cameras and cropped to the object, without part masks, joints or part count, and run its released weights with its default inference settings. We test three inputs: two single frames, one at $t=1$ and the rest frame closest to it in viewpoint (v2, the setting it is trained on); the fused rest frames and the fused last four frames (f2); and four single frames at $t=0$, about $\tfrac13$, about $\tfrac23$ and $1$ (v4). FAMOS does not predict the joint value, so we fit the motion at $t=1$ along its predicted axis as the value that best carries the part's first-state points onto its last-state points. It outputs no surface, so CD-s and CD-m do not apply, and its parts are paired with the true parts on its labeled points. Otherwise scoring follows Sec.~\ref{sec:supp_metrics}. Of the 100 objects, 80 belong to its PartNet-Mobility training split, so Table~\ref{tab:supp_famos} also reports the 20 objects outside it. FAMOS recovers most axes roughly (median error about $1.5^\circ$ against $0.34^\circ$ for \ours), but it misses parts on 17 to 22 objects, which then take the per-part penalty, and its axis errors grow on the objects it was not trained on.

\begin{table}[t]
\centering
\caption{\textbf{Comparison with FAMOS}, a feed-forward articulation model, under the protocol of Table~\ref{tab:main}. Top: all 100 objects. Bottom: the 20 objects outside its training split.}
\label{tab:supp_famos}
\setlength{\tabcolsep}{4pt}
\resizebox{\linewidth}{!}{
\begin{tabular}{lccccc}
\toprule
Method & \makecell{Axis\\Ang $\downarrow$} & \makecell{Axis\\Pos $\downarrow$} & \makecell{Part\\Motion $\downarrow$} & \makecell{CD-s\\$\downarrow$} & \makecell{CD-m\\$\downarrow$} \\
\midrule
\multicolumn{6}{l}{\emph{All 100 objects}} \\
FAMOS v2 & 18.37 & 1.558 & 13.56 & -- & -- \\
FAMOS f2 & 17.34 & 1.369 & 14.09 & -- & -- \\
FAMOS v4 & 21.81 & 1.754 & 14.37 & -- & -- \\
\rowcolor{ourscolor} \ours & \textbf{3.30} & \textbf{0.036} & \textbf{0.35} & 20.11 & 14.64 \\
\midrule
\multicolumn{6}{l}{\emph{20 objects outside the FAMOS training split}} \\
FAMOS v2 & 34.59 & 2.221 & 18.26 & -- & -- \\
FAMOS f2 & 24.34 & 1.529 & 11.94 & -- & -- \\
FAMOS v4 & 35.09 & 2.032 & 17.86 & -- & -- \\
\rowcolor{ourscolor} \ours & \textbf{2.26} & \textbf{0.014} & \textbf{0.19} & 16.94 & 13.71 \\
\bottomrule
\end{tabular}}
\end{table}

\textbf{Language fields.} We also run two released language fields on the same videos. Occam's LGS~\cite{occamlgs} lifts the SAM and CLIP features of LangSplat~\cite{langsplat} onto a trained 3D Gaussian scene. A static field cannot follow a moving object, so we train two static scenes, initialized from the video's depth: one from every other rest frame, and one from the frames with $t\ge0.75$, a window near the other end state that gives the scene enough views but still contains some motion. 4D-LangSplat~\cite{4dlangsplat} is trained on the whole video as one dynamic scene, with $t$ as its time and the rest-frame depth as its initial points, using its released code, its own masks, captions and autoencoders, and its iteration counts. Both receive no part masks and no captions from us. LSE-A and LSP-A use the protocol of Sec.~\ref{sec:supp_metrics}, where each pixel takes the candidate name of highest relevancy against LERF's~\cite{lerf} canonical negatives, and the candidate names are the GPT-4o names our method uses. For State Acc, the feature of each visible part, averaged over its pixels, is compared with the embeddings of the part's five captions, the texts of our caption bank, in each field's own text space: the scene of the nearer end state for Occam's LGS and the time-sensitive field at the frame's $t$ for 4D-LangSplat. Neither field has joints, so LM-A does not apply. Table~\ref{tab:supp_langfields} shows that both name parts less accurately than ArtLang and \ours, and that spatial names are almost never recovered. The two systems differ from ours in more than the state parameterization, since they use their own masks, captions and CLIP text space, so the table compares complete systems, each in its own setup; the effect of state conditioning alone is measured by the ablation of Table~\ref{tab:ablation}. Occam's LGS can only distinguish its two scenes. The time-sensitive feature of 4D-LangSplat changes little with $t$: in every state bucket, the caption it selects is on average the halfway one, so like the static-language variant of Table~\ref{tab:ablation} ($0.141$) it cannot read the state from the feature.

\begin{table}[t]
\centering
\caption{\textbf{Comparison with language fields} on all 100 objects (LSP-A on the 64 objects with a spatial word in their names). ArtLang and \ours as in Table~\ref{tab:main}.}
\label{tab:supp_langfields}
\small
\setlength{\tabcolsep}{5pt}
\begin{tabular}{lccc}
\toprule
Method & LSE-A $\uparrow$ & LSP-A $\uparrow$ & State Acc $\uparrow$ \\
\midrule
Occam's LGS, per state & 0.407 & 0.076 & 0.329 \\
4D-LangSplat & 0.335 & 0.066 & 0.136 \\
ArtLang & 0.897 & 0.701 & -- \\
\rowcolor{ourscolor} \ours & \textbf{0.904} & \textbf{0.735} & \textbf{0.975} \\
\bottomrule
\end{tabular}
\end{table}

\begin{figure*}[t]
\centering
\includegraphics[width=0.9\textwidth]{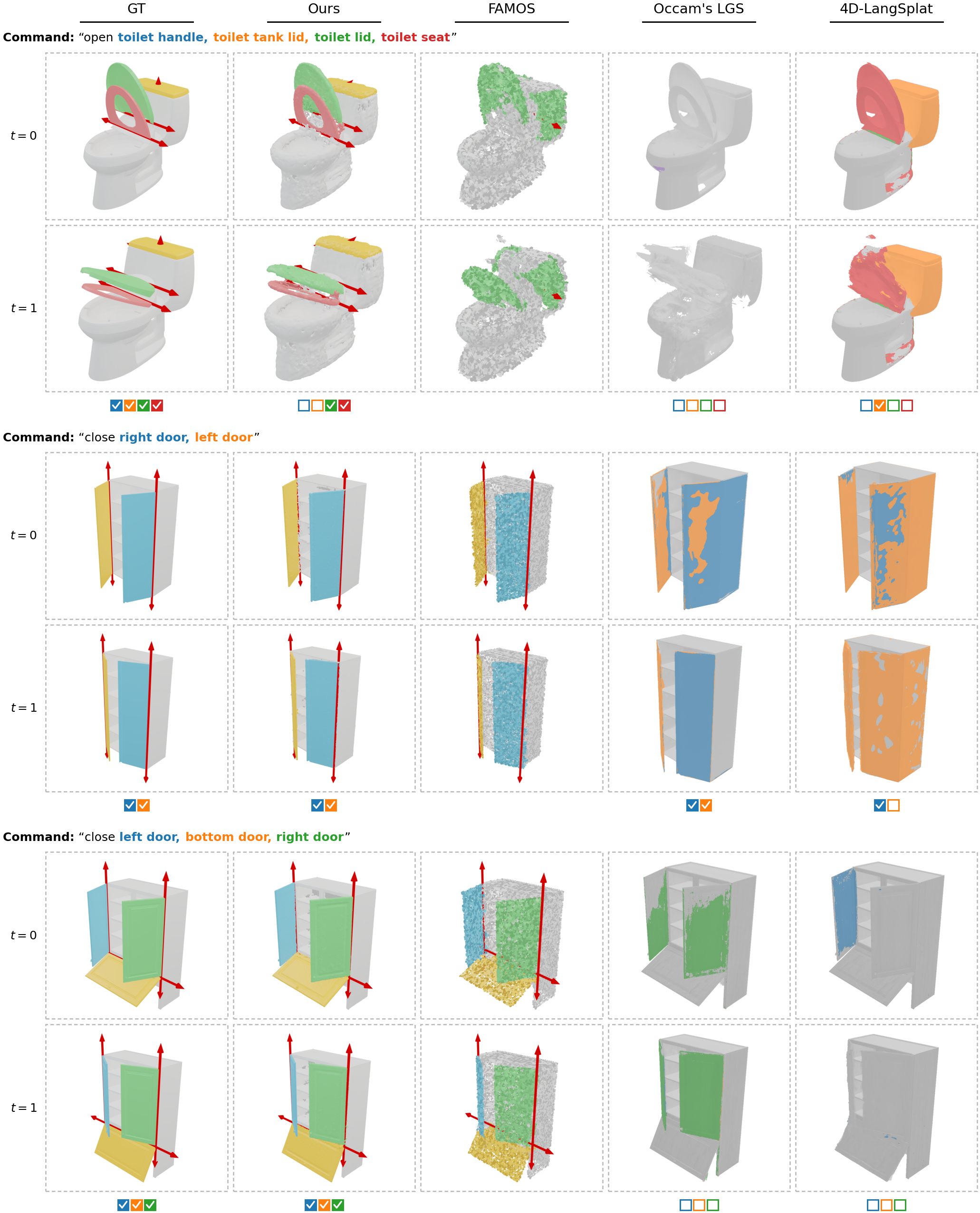}
\caption{\textbf{Comparison with the added baselines (1/2),} the objects of Figure~\ref{fig:supp_qual}. Both end states; FAMOS as its labeled points with its joint axes, Occam's LGS and 4D-LangSplat as their Gaussians colored by the part name the field gives each pixel (no joints). Parts are colored by the true part they are paired with, as in Figure~\ref{fig:supp_qual}. A filled box under a method marks a part that the method gives the name it has in the command.}
\label{fig:supp_qual_nb}
\end{figure*}

\begin{figure*}[t]
\centering
\includegraphics[width=\textwidth]{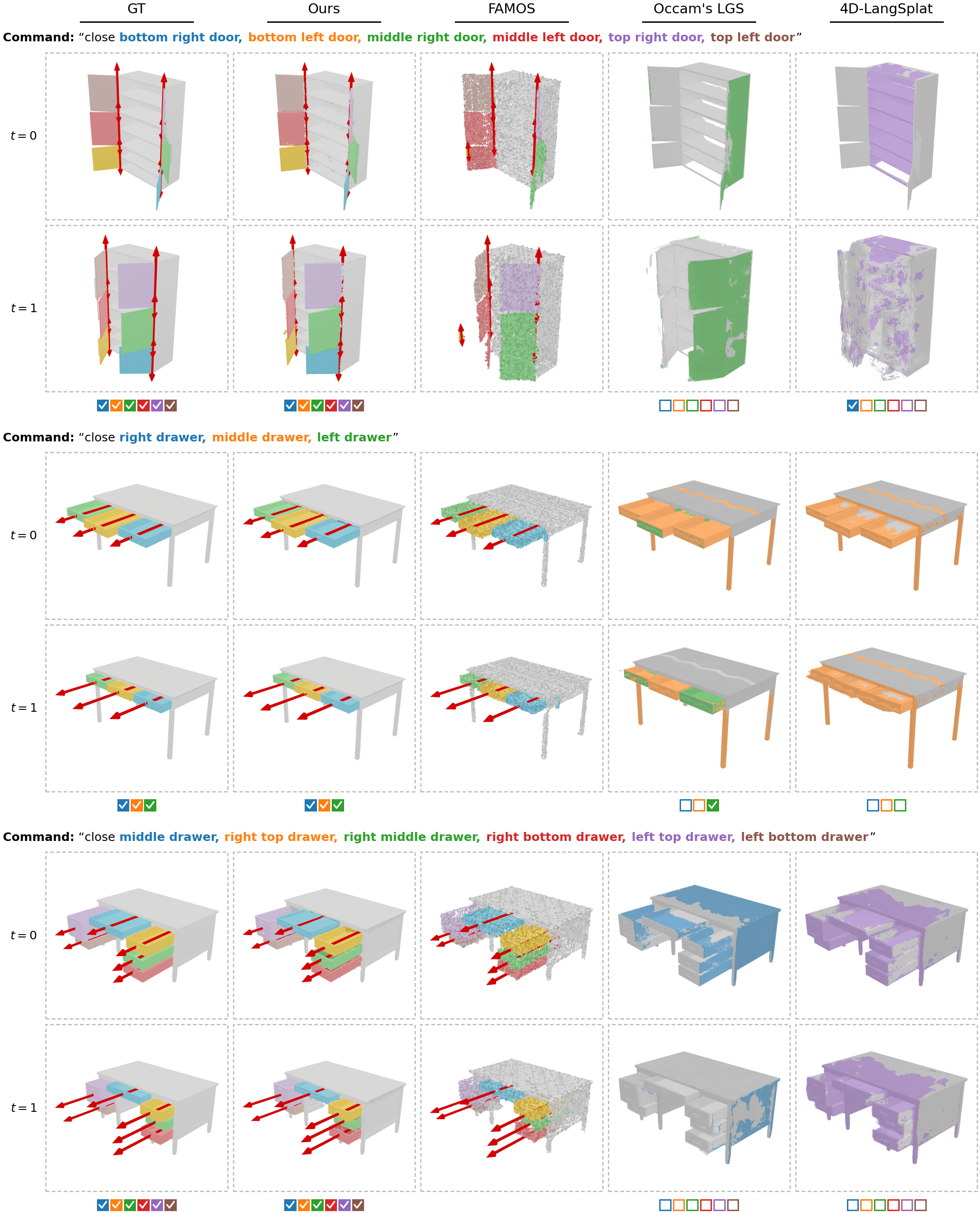}
\caption{\textbf{Comparison with the added baselines (2/2),} the objects of Figure~\ref{fig:supp_qual_2}, as in Figure~\ref{fig:supp_qual_nb}.}
\label{fig:supp_qual_nb_2}
\end{figure*}

\textbf{Qualitative comparison.} Figures~\ref{fig:supp_qual_nb} and~\ref{fig:supp_qual_nb_2} show the objects of Figures~\ref{fig:supp_qual} and~\ref{fig:supp_qual_2} with the added baselines, each in the form of its own output. FAMOS is drawn as the observed points it labels, with its joints. Occam's LGS and 4D-LangSplat are drawn as their Gaussians, rendered with their own renderers from the same camera, every pixel colored by the part name the field gives it under the rule of LSE-A; they predict no joints. FAMOS recovers the large doors and drawers but merges or misses smaller parts. The two language fields cover most of an object with one or two names, so the doors and drawers that a command names are rarely told apart.

\section{Language Stage Settings}
\label{sec:supp_lang_settings}

To test how much State Acc depends on the settings of the language stage, we keep the geometry, the parts and the joints of every object fixed and rerun only the language stage with one setting changed. Table~\ref{tab:supp_langsettings} reports State Acc on all 100 objects. The language dimension $d=48$ and three or seven state buckets stay within $0.015$ of the reported value, so the result does not hinge on these choices. The bucket count changes the states at which a part is captioned, not the five-step scales of Table~\ref{tab:supp_phrases}: with $B=3$ the buckets receive phrases 0, 2 and 4, and with $B=7$ they receive phrases 0, 1, 1, 2, 3, 3 and 4, so two pairs of neighboring buckets share a caption and count as the same state. The chance level of $B=7$ under this rule is $11/49\approx0.22$.

\begin{table}[t]
\centering
\caption{\textbf{Language stage settings.} Only the language stage is rerun, on the same geometry and joints, with one setting changed. All 100 objects.}
\label{tab:supp_langsettings}
\small
\begin{tabular}{lcc}
\toprule
Setting & Chance & State Acc $\uparrow$ \\
\midrule
Default ($d=24$, $B=5$) & 0.20 & 0.975 \\
$d=48$ & 0.20 & 0.987 \\
$B=3$ & 0.33 & 0.989 \\
$B=7$ (five phrases) & 0.22 & 0.982 \\
\bottomrule
\end{tabular}
\end{table}

\section{Parts Moving One After Another}
\label{sec:supp_async}

In the videos of Articulate-100 all parts move together, so one state $t$ describes every frame. To show that the language stage does not rely on this, we render all 100 objects again so that the movable parts move one after another, each in its own window of the video while the others hold still, and give every frame one state per part. The language stage is trained with the per-part states, on the same geometry, joints and caption bank; the reconstruction stages are not rerun. As a reference we train, with identical supervision, a field conditioned on the normalized video time instead of the articulation state. Table~\ref{tab:supp_async} reports State Acc and the query test of Sec.~\ref{sec:supp_results}, where every caption is asked as a free query and counts as correct only when both the part and the state are right. Our field keeps State Acc at the level of Table~\ref{tab:main}. On queries it answers more captions correctly than the time-conditioned field. The time-conditioned field is searched over all 172 frames of the video, a finer grid than the 11 states searched for ours. It returns a moment of the video, and executing that moment sets every part to its state at that frame, while ours sets only the named part and leaves the others where they are (at rest in this test). The \emph{Other parts} column therefore reflects how a query is executed more than the features themselves.

\begin{table}[t]
\centering
\caption{\textbf{Parts moving one after another,} all 100 objects, queries as in Sec.~\ref{sec:supp_results}. \emph{Part and state}: both right. \emph{Within one state}: the part is right and the state is at most one bucket off. \emph{Other parts}: the mean state of the parts not named in the query after it is executed from the rest configuration, as a fraction of their full motion (0 = at rest).}
\label{tab:supp_async}
\small
\setlength{\tabcolsep}{3pt}
\begin{tabular}{lcccc}
\toprule
Conditioned on & \makecell{State\\Acc $\uparrow$} & \makecell{Part and\\state $\uparrow$} & \makecell{Within\\one state $\uparrow$} & \makecell{Other\\parts $\downarrow$} \\
\midrule
Video time & 0.967 & 0.449 & 0.666 & 0.211 \\
\rowcolor{ourscolor} Art.\ state (\ours) & \textbf{0.976} & \textbf{0.568} & \textbf{0.777} & \textbf{0.000} \\
\bottomrule
\end{tabular}
\end{table}

\end{document}